\documentclass{vgtc}

\ifpdf
  \pdfoutput=1\relax                   
  \usepackage{graphicx}                
  \DeclareGraphicsExtensions{.pdf,.png,.jpg,.jpeg} 
\else
  \usepackage{graphicx}                
  \DeclareGraphicsExtensions{.eps}     
\fi%

\graphicspath{{figures/}{pictures/}{images/}{./}} 

\usepackage{microtype}                 
\PassOptionsToPackage{warn}{textcomp}  
\usepackage{textcomp}                  
\usepackage{mathptmx}                  
\usepackage{times}                     
\usepackage{cite}                      
\usepackage{tabu}                      
\usepackage{booktabs}                  
\usepackage{amsmath}
\usepackage{amssymb}
\usepackage{bm}
\usepackage{color}
\usepackage{enumitem}
\usepackage{multirow}

\makeatletter
\def\itemautorefname{\@gobble}
\makeatother

\usepackage{titlesec}
\usepackage[titletoc,toc,title]{appendix}

\usepackage{todonotes}
\newcommand{\Black}[1]{\textcolor[rgb]{0.00,0.00,0.00}{#1}}

\newcommand{\revise}[1]{\Black{#1}}

\onlineid{159}

\vgtccategory{Research}
\vgtcpapertype{application/design study}

\vgtcinsertpkg

\title{Understanding Hidden Memories of Recurrent Neural Networks}

\author{Yao Ming \thanks{e-mail: \{ymingaa, scaoad, rzhangav, zhen, ychench, yqsong, huamin\}@ust.hk} \\
\and Shaozu Cao \footnotemark[1]
\and Ruixiang Zhang \footnotemark[1]
\and Zhen Li \footnotemark[1]
\and Yuanzhe Chen \footnotemark[1]
\and Yangqiu Song, \textit{Member, IEEE} \footnotemark[1]
\and Huamin Qu, \textit{Member, IEEE} \footnotemark[1]
}
\affiliation{\scriptsize Hong Kong University of Science and Technology}

\shortauthortitle{Ming \MakeLowercase{\textit{et al.}}: Visualizing Hidden Memories of Recurrent Neural Networks}

\abstract{Recurrent neural networks (RNNs) have been successfully applied to various natural language processing (NLP) tasks and achieved better results than conventional methods. However, the lack of understanding of the mechanisms behind their effectiveness limits further improvements on their architectures. In this paper, we present a visual analytics method for understanding and comparing RNN models for NLP tasks. We propose a technique to explain the function of individual hidden state units based on their expected response to input texts. We then co-cluster hidden state units and words based on the expected response and visualize co-clustering results as memory chips and word clouds to provide more structured knowledge on RNNs' hidden states. We also propose a glyph-based sequence visualization based on aggregate information to analyze the behavior of an RNN's hidden state at \revise{the} sentence-level. The usability and effectiveness of our method \revise{are} demonstrated through case studies and reviews from domain experts.
} 

\keywords{recurrent neural networks, visual analytics, understanding neural model, co-clustering}

\teaser{
  \centering
  \includegraphics[width=\linewidth]{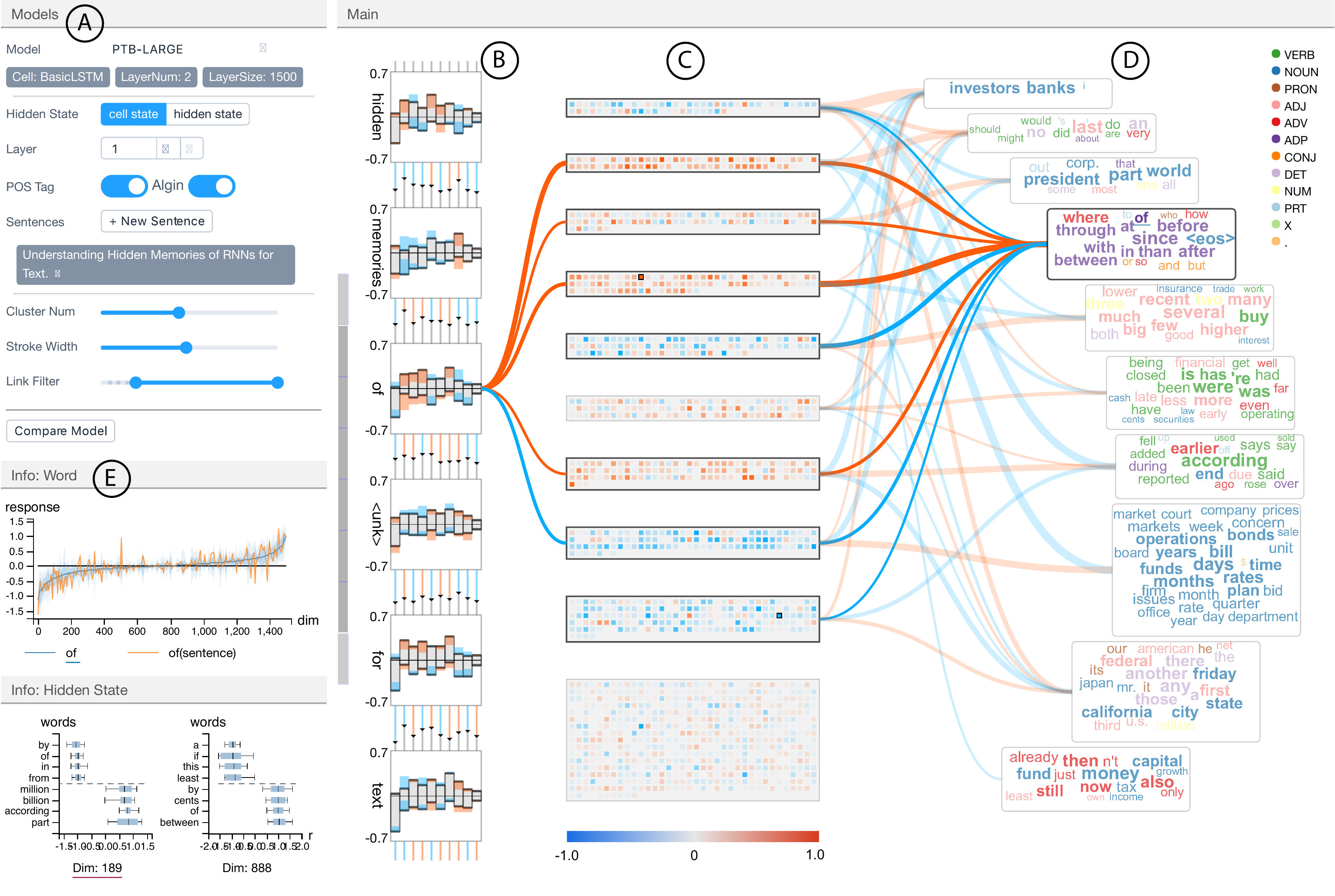}
  \caption{The interface of RNNVis. The control panel (A) shows parameter settings of an RNN and allows users to adjust visualization style. The main view (B-D) contains glyph-based sentence visualization (B), memory chips visualization for hidden state clusters (C), and word clouds visualization for word clusters (D). The detail view (E) shows the distributions of models responses to selected word ``of'' and interpretations of selected hidden units.}
	\label{fig:teaser}
}

\begin{document}


\firstsection{Introduction}

\maketitle
Recurrent neural networks (RNNs) are deep learning models that are very effective in modeling sequence data such as text and speech. RNNs and their variants, especially long short term memory (LSTM) networks and gated recurrent units (GRU), have been successfully applied in various natural language processing (NLP) applications, which include language modeling \cite{mikolov10lm}, machine translation \cite{sutskever14seq2seq}, speech recognition \cite{graves13speech} and sentiment analysis \cite{tang15sentiment}. Recent work extended RNNs to solve challenging tasks such as reading comprehension \cite{karlmoritz15comprehend} and video analysis \cite{donahue15video}, and achieved better performance than traditional approaches.

Despite their impressive performances, RNNs are still ``black boxes'' that are difficult for humans to understand. As a basic component for various applications, RNNs have recently attracted \revise{researchers' interests} in designing better model architectures (e.g., the recently proposed GRU \cite{cho14gru}). However, the lack of understanding of how RNN models work internally with their memories has limited researchers' ability to introduce further improvements. \revise{A recent study \cite{kdd16trust} also emphasized the importance of interpretability of machine learning models in building users' trust: \textit{if users do not trust the model, they will not use it.}}

The three major challenges of interpreting RNNs are as follows. First, RNNs maintain memory-like arrays called hidden states which store information extracted \revise{from} a long input sequence (e.g., texts or audios). As the input \revise{comes} in, hundreds or thousands of hidden state units are updated with nonlinear functions specified by millions of parameters. The size of hidden states and parameters \revise{brings} difficulties of analyzing an RNN's behavior. The number of parameters will further increase as the model becomes more complex. Second, RNNs mainly deal with sequential data such as texts and audios. The complex sequential rules (e.g., grammar, language models and social language codes) embedded in texts are intrinsically difficult to be interpreted and analyzed. Third, through experiments, we found that semantic information in hidden states \revise{is} highly distributed, i.e., each input word will generally \revise{result} in changes almost every hidden state units. Similarly, each hidden state unit may be highly responsive to a set of words. The many-to-many relationship between hidden state units and words further impedes researchers from understanding the information embedded in hidden states of RNNs.

\revise{To address these challenges in interpreting RNN models, a} few recent studies have compared \revise{the} behavior of hidden states in different RNN architectures. Karpathy et al. \cite{DBLP:journals/corr/KarpathyJL15} found that \revise{some} hidden state units can be interpreted as language context, such as quotes or punctuations, whereas others do not seem to represent any semantics. Strobelt et al. \cite{DBLP:journals/corr/StrobeltGHPR16} used parallel coordinates to present values of a few selected hidden state units with a pattern searching function for experts to explore the dataset. However, most existing studies focused on a limited aspect of the model only. There \revise{is still a lack of} thorough understanding of what information is recorded \revise{in} hidden states and how the information is distributed and updated.

In this work, we develop a general visual analytics system called RNNVis (\autoref{fig:teaser}) for deep learning practitioners to understand and diagnose RNNs and explore the behavior of their hidden memories. 
Inspired by previous \revise{work} in convolutional neural networks (CNNs), we introduce a technique to interpret functions of each hidden state unit using its highly correlated words from the input space. The correlation between a hidden state unit and a word is measured using the expected \revise{update} of the hidden state unit given the word as input.
To understand how \revise{the} functions of hidden state units in an RNN are composed together, we model the relation between hidden states and words as a bipartite graph, in which hidden state units and words are treated as two types of nodes connected by weighted edges. 
To enable scalable visual analysis for RNNs with \revise{a} large size of hidden states, \revise{we arrange the bipartite graph} into co-clusters. Hidden state clusters and word clusters are then visualized as memory chips and word clouds, which are convenient for explorations.
Based on the co-cluster visualization, \revise{the sentence-level behavior} of an RNN \revise{is} examined by a glyph-based visualization using aggregate statistics of hidden states. Rich interactions are also provided for users to explore and compare the hidden behavior of different models.

Our major contributions are:
\begin{itemize}
\setlength\itemsep{-0.5em}
\item The design and implementation of RNNVis, a visual analytics system for understanding, comparing, and diagnosing RNNs for general text-based NLP tasks.
\item A new technique for bridging hidden states and textual information with expected response.
\item A glyph-based design for sequence visualization to analyze the sentence-level behavior of RNNs.
\item Case studies on the analysis of hidden mechanisms of RNNs with different architectures using the proposed system.
\end{itemize}

\section{Related Work}
In this section, we review some related work on understanding RNNs and approaches to visualize neural networks.

\subsection{Understanding RNNs}
In the field of computer vision, significant efforts were exerted to visualize and understand how the components of a CNN work together to perform classifications. These studies (Zeiler $\&$ Fergus \cite{DBLP:conf/eccv/ZeilerF14}, Dosovitskiy $\&$ Brox \cite{DBLP:conf/cvpr/DosovitskiyB16}) provided researchers with insights of neurons' learned features and inspired designs of better network architectures (e.g., the state-of-the-art performance on the ImageNet benchmark in 2013 proposed by Zeiler $\&$ Fergus \cite{DBLP:conf/eccv/ZeilerF14}).

However, these methods are difficult to be applied or adapted to RNNs. The lack of interpretability of the hidden behavior in RNNs recently raised concerns in the NLP community. As a critical basis of various neural models in NLP, RNN has attracted an increasing number of studies that focus on analyzing its hidden activities and improving its architectures. 
These studies can be divided into two categories: performance-based analysis and interpretability-guided extensions.

Performance-based methods analyze model architectures by altering critical network components and examining the relative performance changes. Greff et al. \cite{DBLP:journals/corr/GreffSKSS15} conducted a comprehensive study of LSTM components. Chung et al. \cite{DBLP:journals/corr/ChungGCB14} evaluated the performance difference between GRUs and LSTMs. Jozefowicz et al. \cite{DBLP:conf/icml/JozefowiczZS15} conducted an \revise{automatic} search among thousands of RNN architectures. These approaches, however, only show overall performance differences regarding certain architectural components, and provide little understanding \revise{of} the contribution of inner mechanisms.

Another worth mentioned type of neural models extends RNN with an attention mechanism to improve \revise{the} performance \revise{on} specific tasks. Bahdansu et al. \cite{DBLP:journals/corr/BahdanauCB14} applied the attention in machine translation and showed the relationship between source and target sentences. Xu et al. \cite{DBLP:conf/icml/XuBKCCSZB15} \revise{designed} two attention-based models in image captioning, which revealed the reasons behind the effectiveness of their models. Although the attention mechanism can benefit the interpretation without extra effort, it requires jointly training different models or modifying the original model, which limits its application in general RNN models.

\subsection{Visualization for Machine Learning}
There is a trend for combining visualization and machine learning in recent years. 

\revise{On the one hand, visualization has been increasingly adopted by the machine learning community to analyze\cite{squares17}, debug \cite{visualdebug16}, and present \cite{tensorflow16} machine learning models. On the other hand, a number of human-in-the-loop methods have been proposed as competitive replacements of full-automatic machine learning methods. These methods include: visual classification\cite{conf/sdm/TeohM03,conf/ieeevast/ChooLKP10, 10.1109/TVCG.2011.212}, visual optimization\cite{kapoor2010interactive,talbot2009ensemblematrix}, and visual feature engineering \cite{Lehmann:2017:cgfc,FeatureInsight15,DBLP:journals/corr/SinghP14}.}

\revise{In the field of deep learning, some recent studies have utilized visualization to help understand RNNs.}
Tang et al. \cite{DBLP:journals/corr/TangSWFZ16} studied the behavior of LSTM and GRU in speech recognition by projecting sequence history. Karpathy et al. \cite{DBLP:journals/corr/KarpathyJL15} showed that certain cell states can track long-range dependencies by overlaying heat map on texts. Li et al. \cite{LiCHJ16} also used heat maps to examine sensitiveness of different RNNs to words in a sentence. However, their visualizations only provided \revise{an} overall analysis of RNNs. These studies did not explore RNN's hidden states in detail. 

In the field of visualization, recent \revise{work has} exhibited the effectiveness of visual analytics in understanding, diagnosing and presenting neural networks. Liu et al. \cite{DBLP:journals/tvcg/LiuSLLZL17} treated deep CNN as a directed acyclic graph and built an interactive visual analytics system to analyze CNN models. Rauber et al. \cite{DBLP:journals/tvcg/RauberFFT17} applied dimensionality reduction to visualize learned representations, as well as the relationships among artificial neurons, and provided insightful visual feedback of artificial neural networks. \revise{While} visualization has achieved considerable success on CNNs, \revise{little} work has focused on RNNs. Most related to our work, Strobelt et al. \cite{DBLP:journals/corr/StrobeltGHPR16} has proposed an interactive visualization system to explore hidden state patterns similar to a given phrase on a dataset. This system also allows users to flexibly explore given dimensions of hidden states. However, the parallel coordinates design is not scalable for efficiently analyzing hundreds or thousands of hidden state dimensions.

\subsection{Co-Clustering and Comparative Visualization}
Our work is closely related to two techniques: co-clustering and comparative visualization. 

We formulate the relation between hidden state units and discrete inputs of RNNs as bipartite graphs to investigate the structure of information stored hidden states.
Co-clustering is a widely used method for analyzing bipartite graphs, which simultaneously clusters two kinds of entities in a graph  \cite{cocluster04survey}. 
{Some recent work} combined co-clustering with visualization to assist intelligence analysis, where different types of entities are considered \cite{bixplorer13,bicluster14}.
A most recent work proposed by Xu et al. \cite{xu16pvis} presented an interactive co-clustering visualization where cluster nodes are visualized as adjacency matrices or treemaps. Although both adjacency matrices and treemaps used in this visualization are  well established, none could be adjusted to visualize abstract entities like hidden states. 

Comparative visualization was adopted to fulfill the design requirements of RNNVis. Gleicher et al. \cite{compare11} suggested three typical strategies for comparative visualization, namely, juxtaposition (or separation), superposition (or overlay), and explicit encoding. We mainly employ juxtaposition and superposition for comparing RNNs at three different levels, namely, detail, sentence, and overview levels. The details of the design choices are discussed in \autoref{sec:interaction}.

\section{Backgrounds} \label{sec:bg}


The basic architecture and concepts of RNNs are introduced to serve as a basis for the discussions of the next sections.

\begin{figure}[ht]
 \centering 
 \includegraphics[width=\columnwidth]{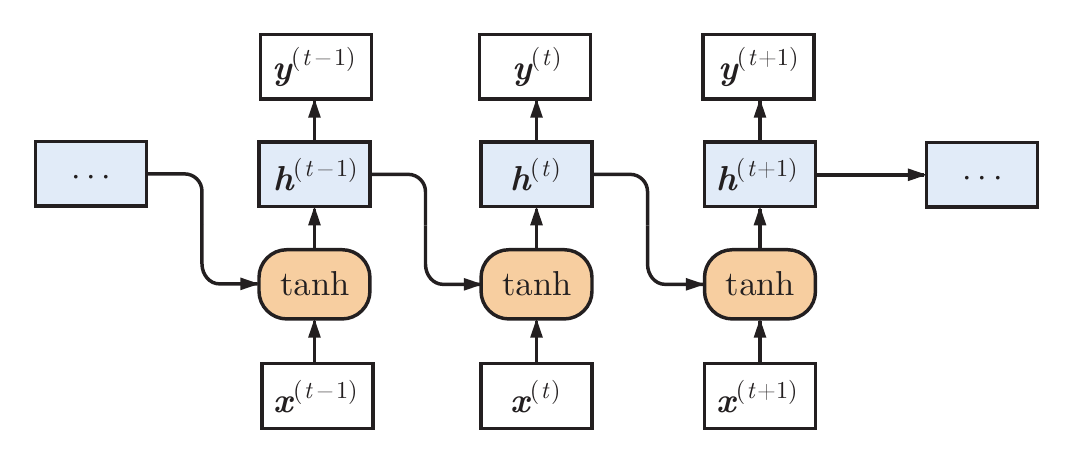}
 \caption{Illustration of the vanilla recurrent neural network.}
 \label{fig:rnn-illustrate}
\end{figure}

\textbf{Recurrent Neural Network.} As shown in \autoref{fig:rnn-illustrate}, a vanilla RNN takes sequential inputs $\{\bm{x}^{(0)}, \cdots, \bm{x}^{(T)}\}$, where $\bm{x}^{(t)} \in \mathbb{R}^m$ and maintains a time-variant \textit{hidden state} vector $\bm{h}^{(t)} \in \mathbb{R}^n$. At step $t$, the model takes input $\bm{x}^{(t)}$, and updates the hidden state $\bm{h}^{(t-1)}$ to $\bm{h}^{(t)}$ using:
\begin{equation}
\label{eq:rnn}
\bm{h}^{(t)} = f(\bm{W} \bm{h}^{(t-1)} + \bm{V} \bm{x}^{(t)})
\end{equation}
where $\bm{W}$ and $\bm{V}$ are \textit{weight matrices} and $f$ is a nonlinear \textit{activation function}. In this paper, we use $\tanh$ as activation function, which constraints value range of $\bm{h}^{(t)}$ to $(-1, 1)$. 

\revise{\textbf{Softmax Output.}}
After the updates at each step, $\bm{h}^{(t)}$ may be further processed or directly used as output. 
\revise{For instance, to perform classification, a probability distribution $p$ over $K$ classes can be computed after processing the whole sequence at step $T$:
\begin{equation}
\label{eq:softmax}
p_i = \text{softmax}(\bm{U}\bm{h}^{(T)})_i = \frac{\exp(\bm{u}_i^T \bm{h}^{(T)})}{\sum_{j=1}^{K} \exp(\bm{u}_j \bm{h}^{(T)})}
\end{equation}
where $\bm{U} = [\bm{u}_1, \cdots, \bm{u}_n]^T$ is the output projection matrix.
}

\revise{We} can associate RNNs with Turing Machines in the sense that both of them maintain a piece of ``memory''. As the input sequence passed in step by step, an RNN updates its memory and outputs some results according to its current memory. 

\revise{In this paper, we will also use two RNN variants: long short-term memory (LSTM) networks and gated recurrent units (GRUs). Their definitions can be found in Appendix A. The multi-layer models are also defined in the appendix.}



\begin{figure}[ht]
 \centering 
 \includegraphics[width=\columnwidth]{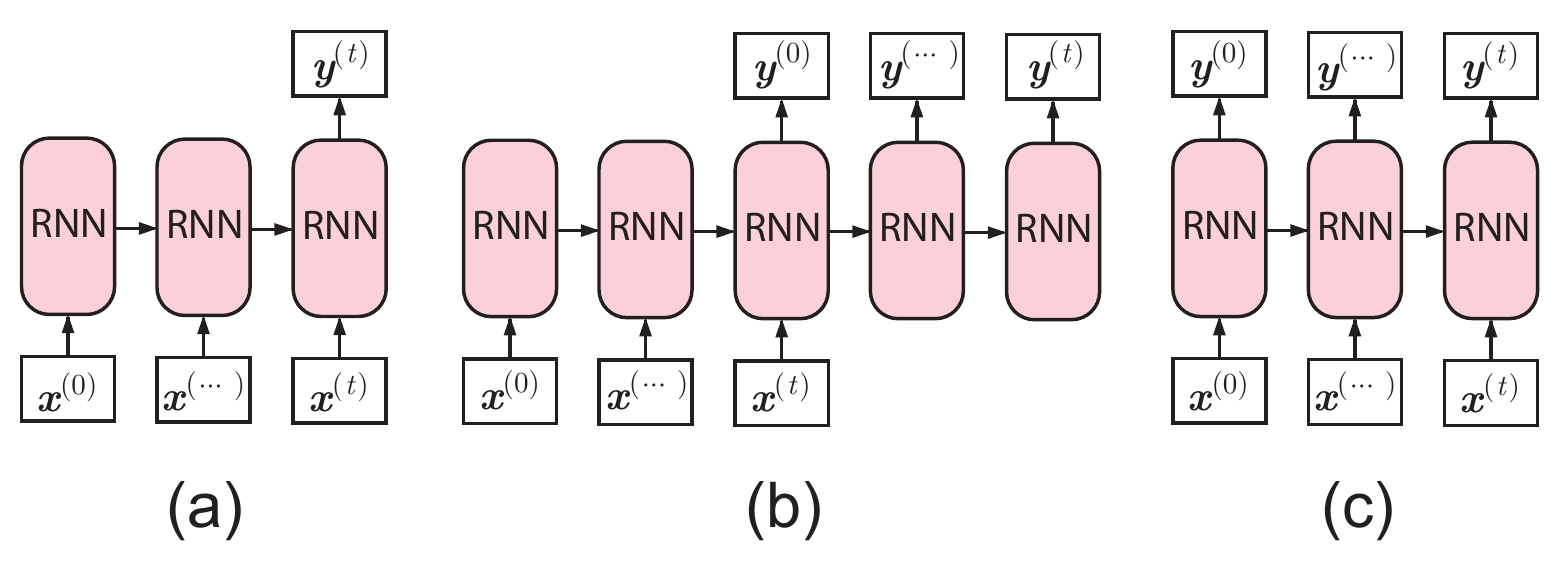}
 \caption{Input-output schemes of RNN.}
 \label{fig:io-scheme}
\end{figure}

\textbf{Input-Output Schemes.} In different application scenarios, an RNN typically has one of the following input-output schemes: sequence-to-one, sequence-to-sequence, synced sequence-to-sequence, as shown in \autoref{fig:io-scheme}. RNNs can take a sequence of inputs and only output at the end of the whole sequence (\autoref{fig:io-scheme}(a)), which is typical in sentiment analysis or document classification. The sequence-to-sequence (\autoref{fig:io-scheme}(b)) formulation is widely used in machine translation (e.g., translating a sentence to another language). More generally, certain tasks (e.g., language modeling and video classifications) generate outputs at each step of the input sequence \revise{(\autoref{fig:io-scheme}(c))}. To evaluate our proposed system, we used sequence-to-one (e.g., sentiment analysis for texts) and synced sequence-to-sequence (language modeling) for illustrations. Other RNN-based models can also be analyzed given that the proposed visual analytic method only requires recorded information of hidden states.

\section{System Design}
In this section, the requirements of the current system for interpreting and analyzing the hidden states of RNNs are discussed and formulated. The proposed techniques and visual designs are presented in \autoref{sec:model} and \autoref{sec:design} respectively.

Throughout the design and implementation of RNNVis, we closely worked with two experts in the deep learning and NLP domains, who are also co-authors of this work. One expert (E1) is a deep learning researcher with strong industrial experience, and the other expert (E2) is a senior researcher \revise{who} specializes in natural language understanding and text mining. Decisions on algorithm design (\autoref{sec:model}) and visual encoding choices (\autoref{sec:design}) for the system are determined through iterative discussions with collaborators. 

\subsection{Requirement Analysis} \label{sec:requirement}

The major focus of RNNVis is to provide intuitive interpretations of RNN's hidden states, and make use of the knowledge to diagnose RNN models and inspire better architecture designs. Based on the discussions with domain experts and literature review, the specific requirements are formulated as follows:

\begin{enumerate}[label=\textbf{R\arabic*}]
\setlength\itemsep{0em}
\item \label{req:interpret}
\textbf{Clearly interpret the information captured by hidden states}.
Current visualization techniques for hidden states either select a few state units that may be easy to interpret \cite{DBLP:journals/corr/KarpathyJL15}, or directly map hidden state values to visualizations like parallel coordinates \cite{DBLP:journals/corr/StrobeltGHPR16}. However, as most interpretable information is distributively stored in hidden states, direct visualization cannot provide \revise{explanations} for each hidden \revise{unit}. Thus, the most basic requirement is to visually explain the semantic information captured by each hidden state unit. For example, what kinds of words or grammars are captured and stored in a hidden unit?

\item \label{req:overview}
\textbf{Provide the overall information distribution in hidden states}. 
Besides the interpretation of individual hidden states, an overview of how \revise{semantic} information \revise{is} structured within hidden memories can provide experts with a full picture of the model. For example, how is the stored information differentiated and correlated across hidden states?

\item \label{req:sequence}
\textbf{Explore hidden states mechanisms at the sequence-level}. 
During discussions, domain experts expressed the need to analyze RNN's ability in modeling sequences, which is the most advantageous feature of RNN. A widely used visualization technique for RNN is to project the learned word embedding to a 2-D space. However, word-level visualization cannot reveal the reasons behind RNN's effectiveness in modeling sequence. That is, how does the internal memory updating mechanism of an RNN result in its specific behavior when dealing with sequences?

\item \label{req:detail}
\textbf{Examine detailed statistics of individual states}. Experts also suggested that concise and detailed information, such as the distribution of hidden state values or gate activations, is required for quantitatively analysis. 

\item \label{req:compare}
\textbf{Compare learning outcome of models}.
A general but important requirement is \revise{the} comparison of different RNN architectures. For example, what \revise{are} the internal reasons that one model is better than the other? What are the mechanisms that arm LSTMs the ability to master long-term dependency?

\end{enumerate}
\begin{figure}[hb]
 \centering 
 \includegraphics[width=0.9\columnwidth]{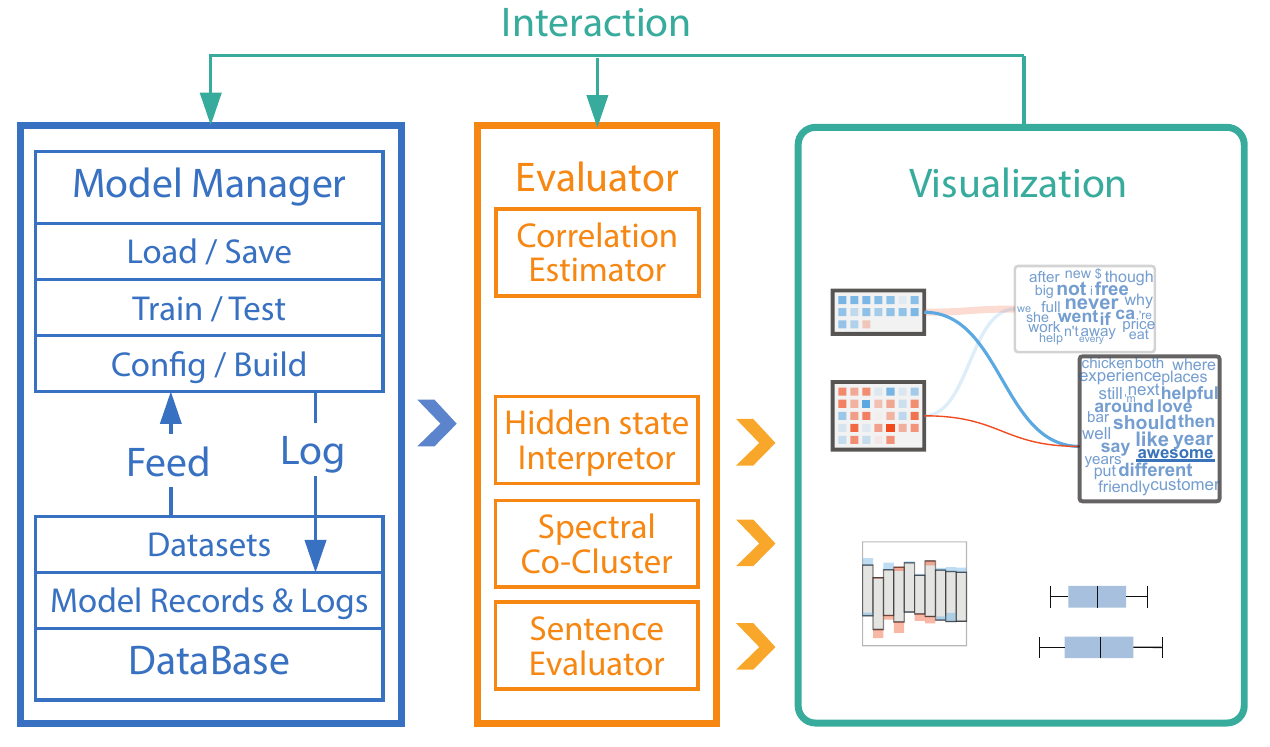}
 \caption{System architecture. Left: the model manager that manages RNN models and datasets. Middle: the RNN evaluator that evaluates the model provided by the model manager. Right: the visualization module.}
 \label{fig:architecture}
\end{figure}
\setlength{\belowcaptionskip}{-10pt}

\subsection{System Overview}

As shown in \autoref{fig:architecture}, RNNVis consists of three major modules: \textit{model manager}, \textit{RNN evaluator}, and \textit{interactive visualization}. 

The model manager utilizes TensorFlow \cite{tensorflow16}, a numerical computation library for machine learning, to build, train and test RNN models. The model manager is designed to loosely couple with other modules to offer flexibility when adapted to other RNN based models or different machine learning platforms. \revise{
Users are allowed to edit configuration files to easily change models' architectures. New datasets can also be added by extending the pre-defined data pipeline provided by the model manager.
}

The RNN evaluator then analyzes trained models to extract learned representations in hidden states, and further processes the evaluation results for visualization. This module also offers functionalities to derive interpretations for each hidden state unit in word space. The relation between hidden states and input words are then stored as a bipartite graph. For scalable visualization, we applied a co-clustering algorithm to simultaneously cluster the hidden state space and word space, so that hundreds of hidden states can be explored with ease.

\revise{Evaluation results of the RNN models are finally provided as a co-clustering visualization (\autoref{fig:teaser}). Users first select a trained model in the control panel (A), where architectural parameters of the model are also listed. The main view (B-D) then presents hidden state clusters as memory chips (C), word clusters as word clouds (D), and layouts them in a way that highly correlated co-clusters are closely positioned. To analyze the sequential behavior of the model, users can utilize sequence visualization (B) through inputting sentences in the control panel. Users can further explore the distribution of model's responses to particular words by clicking words or memory cells in the main view and examine the detail view (E). The control panel also provides controls to help users adjust visualization style. Users can also learn the usage of the system by watching a guidance video which introduces each view.}

\section{RNN Evaluator} \label{sec:model}
Before \revise{presenting the} interface and interaction design, we first discuss the techniques used in the RNN evaluator.

\subsection{Interpreting hidden states}
In CNNs, the features learned by neurons can be visually explained using images derived by activation maximization \cite{activate_max09} or code inversion \cite{DBLP:conf/eccv/ZeilerF14}, since the input space of CNN is continuous. However, the input space of RNNs applied in NLP usually consists of discretized words or characters, where these image-based methods fail to generalize. Although gradient-based methods \cite{LiCHJ16} managed to provide overall interpretations on model's predictions, they are difficult to be applied to intermediate hidden states \revise{$\bm{h}^{(t)}$, whose gradients are very sensitive with regards to $\bm{h}^{(t-1)}$}.

Inspired by the idea of using interpretable representations to explain functions of network components, we propose a method to intuitively interpret individual hidden state unit using words or characters. For simplicity, we only use \revise{the} word-level model \revise{for} illustration, but character-level models can also be used.

\revise{
We first show that the numerator of probability $p_i$ in \autoref{eq:softmax} can be decomposed into a product of factors using the method proposed by Murdoch and Szlam\cite{autorule17} (we always set $\bm{h}^{(0)} = \bm{0}$):
\begin{equation}
\exp(\bm{u}_i^T \bm{h}^{(T)}) = \exp \left (\sum_{t=1}^T \bm{u}_i^T (\bm{h}^{(t)} - \bm{h}^{(t-1)})\right ) = \prod_{t=1}^{T}\exp(\bm{u}_i^T \Delta\bm{h}^{(t)}).
\end{equation}
Here, $\exp(\bm{u}_i^T \Delta\bm{h}^{(t)})$ can be interpreted as the multiplicative contribution of word $t$ to the predicted probability of class $i$, and $\Delta\bm{h}^{(t)} = \bm{h}^{(t)} - \bm{h}^{(t-1)}$ can be regarded as \textit{model's response} to input word $t$.}

\revise{
Although $\bm{h}^{(t)}$ is calculated by a non-linear transformation of $\bm{h}^{(t-1)}$ and $\bm{x}^{(t)}$, $\Delta\bm{h}^{(t)}$ is deterministic to the input $\bm{x}^{(t)}$ when the previous history $\bm{h}^{(t-1)}$ is given. Thus, $\Delta\bm{h}^{(t)}$ can reflect to what degree the model's hidden state is influenced by the input $\bm{x}^{(t)}$. However, given the same word $\bm{x}$, $\Delta \bm{h}^{(t)}$ might vary due to its dependence on $\bm{h}^{(t-1)}$. Consequently, we formulate $\bm{x}^{(t)}$ and $\bm{h}^{(t)}$ as random variables, and use model's expected response to a word $w$ as a more stable measure of the word's importance on hidden state units. We will also show that this formulation is empirically effective.}

\revise{
In NLP, where inputs are sequences, it is common to regard input words $w^{(t)}$ as random variables \cite[Chapter~4]{Daniel09SLP}. Thus, the corresponding word embedding vectors $\bm{x}^{(t)}$ can be regarded as discrete random variables. Since the hidden state $\bm{h}^{(t)}$ is deterministically computed from a sequence of random variables $\bm{x}^{(0)},\cdots,\bm{x}^{(t)}$, it is also eligible to consider $\bm{h}^{(t)}$ as random variables\cite[p.~389]{Goodfellow-et-al-2016}.
}

\revise{
The expected response to a word $w$ is then computed by Adam's Law:
\begin{equation}
\label{eq:expectation}
s(\bm{x}) = E(\Delta\bm{h}^{(t)}\mid \bm{x}) = E(E(\Delta\bm{h}^{(t)}\mid \bm{x}, \bm{h}^{(t-1)}))
\end{equation}
where $\bm{x}$ is the embedding vector of $w$, and $s(\bm{x})_i$ represents the relation between the $i$th hidden state unit $h_i$ and $w$. Note that with the $tanh$ activation function, the response $s(\bm{x})_i$ can have either positive or negative value. A larger absolute value of $s(\bm{x})_i$ indicates that $\bm{x}$ is more ``salient'' or important to the hidden state unit $h_i$.
The advantage of \autoref{eq:expectation} is that, with enough data, we can easily estimate the expected response using all observations of $\Delta\bm{h}^{(t)}$ on the word $w$:
\begin{equation}
\hat{s}(\bm{x}) = \frac{1}{\sum_{\bm{x}^{(t)} = \bm{x}} 1} \sum_{\bm{x}^{(t)} = \bm{x}}{\Delta\bm{h}^{(t)}}.
\end{equation}
The explanation of a hidden state unit $i$ is then formulated as $m$ words with top $m$ absolute expected responses. In our prototype, users can adjust the parameter $m$.
}


\revise{
For an LSTM, which has two state vectors, we calculate the model's expected response based on the update of cell states, i.e., $s(\bm{x}) = E(\Delta\bm{c}^{(t)}\mid \bm{x})$. The reason of this specification is that cell state $\bm{c}^{(t)}$ is considered to maintain \revise{long-term} memory, while $\bm{h}^{(t)}$ is directly computed from cell state and used for output. However, based on the above formulation, RNNVis can also be used to analyze the behavior of $\bm{h}^{(t)}$ of an LSTM.
}

\begin{figure}[ht]
 \centering 
 \includegraphics[width=\columnwidth]{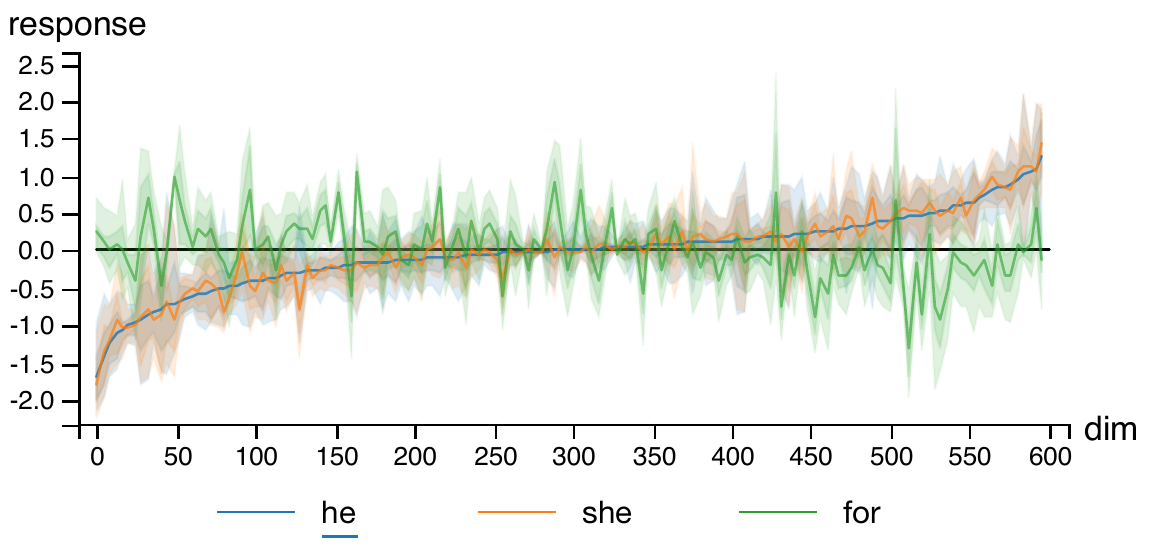}
 \caption{Distribution of \revise{an LSTM's response} to input words ``he'' and ``she'' and ``for''. Horizontal axis represents 600 dimensions of the hidden states of the last layer. \revise{The dimensions are sorted according to the model's estimated expected response to ``he'', $\hat{s}(``he")$.} The darker area denotes 25\% to 75\% range of the distribution, while the lighter area denotes 9\% to 91\% range.}
 \label{fig:he-she-for}
\end{figure}

\revise{In \autoref{fig:he-she-for}, we show the distributions of a two-layer LSTM's responses, $\Delta\bm{c}^{(t)}$, given three different words. The LSTM has 600 cell state units per layer and is trained on the Penn Tree Bank (PTB) dataset \cite{ptb93}.} We can see that the hidden state units in the left and right end are highly responsive to ``he'' and ``she''. In addition, we can see that the model's response patterns differ a lot between prepositions (``for'') and pronouns (``he'' and ``she'').

\subsection{Co-clustering Hidden States and Words} \label{sec:coclst}
In NLP, state-of-the-art RNN models have about two to four layers, with the size of hidden states vector per layer ranges from several \revise{hundred} to a few \revise{thousand}. Showing the interpretations of one hidden state unit at a time creates cognitive \revise{burdens} to users, and does not offer clues about RNN's high-level behavior of all hidden states as a whole. 

To generate an overview for hidden \revise{units or neurons, the} machine learning community has widely adopted projection methods (e.g., t-SNE and PCA) to explore or illustrate the patterns inside model's learned features \cite{visproject16}. For example, in NLP, projection is a common approach in demonstrating model's effectiveness in mastering \revise{word-level} semantic relationships such as analogy. However, \revise{projecting} word embedding alone only provides an overall structure of the learned representations. Analyzing a large number of hidden states still requires a more concise overview. 

\revise{Given the expected response $s(\bm{x})$ derived from \autoref{eq:expectation}, for each word, we have the expected responses of $n$ hidden units, and for each hidden unit, we have its expected responses to $N$ words. By viewing words and hidden units as nodes, we can model this multi-to-multi relation as a bipartite graph $G = (V_w, V_h, E)$, where $V_w$ and $V_h$ are sets of word nodes and hidden unit nodes, respectively.} The relations between hidden state units and words are then treated as weighted edges \revise{$E = \{ e_{i, j} = s(\bm{x}(w_i))_j \mid w_i \in V_w, h_j \in V_h \}$, where $\bm{x}(w_i)$ is the embedding vector of word $w_i$, and $h_j$ is the $j$th hidden unit of hidden state vector $\bm{h}$. As a natural way of analyzing bi-graphs, co-clustering (or bipartite graph partitioning)}\cite{xu16pvis} is used to structure hidden state space and word space for easier exploration while preserving the relation between two spaces. In the prototype of RNN evaluator, the spectral co-clustering algorithm \cite{coclustering01} with k-means++ initialization is used to simultaneously cluster hidden state units and words. Other co-clustering algorithms can also be applied. 

In the presented case, structured hidden state clusters and word clusters also facilitate the provision of overview visualization as required in \autoref{req:overview}. \revise{Our method outperforms projection-based methods in that it shows the word-level semantics directly with clusters. For instance, the highlighted word cluster (in the form of a word cloud) in \autoref{fig:teaser}D roughly presents ``preposition''.} In addition, the semantics formed by word clusters \revise{also serves} as concise interpretations for corresponding hidden state clusters.

\subsection{Sequence Analysis} \label{sec:model-sequence}

To support sequence-level analysis as required in \autoref{req:sequence}, we discussed with the experts and formulated a few aggregate measurements to profile RNN's behavior at sequence-level. All the measurements are defined as cluster-level summaries at each step of $p$ hidden state clusters $\{H_1, H_2, \cdots, H_p\}$ obtained from \autoref{sec:coclst}. We will show the usefulness of these measurements in \autoref{sec:design-seq}.

\revise{\textbf{Aggregate Information} of a cluster $H_i$ at a step $t$ is defined as $\alpha_i^{(t)} = (\alpha_{i+}^{(t)}, \alpha_{i-}^{(t)})$, where $\alpha_{i+}^{(t)}$ and $\alpha_{i-}^{(t)}$ are the sums of positive and negative hidden units in cluster $H_i$:
\begin{equation}
\alpha_{i+}^{(t)} = \sum_{h_j \in H_i, h_j > 0} {h_j^{(t)}},\qquad \alpha_{i-}^{(t)} = \sum_{h_j \in H_i, h_j < 0} {h_j^{(t)}}.
\end{equation}
Since a larger absolute value of a hidden unit $|h_j|$ represents that it is more activated or stores more information, $\alpha_{i+}$ and $\alpha_{i-}$ represent to what degree a cluster of hidden units is positively or negatively activated.
We separate the information into positive $\alpha_{i+}$ and negative $\alpha_{i-}$ because they generally refer to different semantics. For example, in sentiment analysis, if a hidden unit encodes positive sentiment with positive value, it will very likely encode negative sentiment with negative value. Besides, 
the sign of $\alpha_{i+} + \alpha_{i-}$ can flag the characteristics (positive or negative) of the information stored in cluster $H_i$.
}

\revise{\textbf{Updated information} of $H_i$ is defined as $\delta^{(t)} = (\Delta \alpha_{i+}^{(t)}, \Delta \alpha_{i-}^{(t)} )$, where $\Delta \alpha_{i+}^{(t)} = \alpha_{i+}^{(t)} - \alpha_{i+}^{(t-1)}$} is the change of positive aggregate information of cluster $H_i$ at step $t$, whereas $\Delta \alpha_{i-}^{(t)}$ is the change of negative information. 
\revise{The absolute updated information $|\Delta \alpha_{i+}^{(t)} + \Delta \alpha_{i-}^{(t)}|$} can be used to measure the sensitiveness of hidden state cluster $H_i$ with the current input. 
\revise{This measure is useful for examining the most responsive hidden state clusters to the current input. Particularly, a high value of $|\Delta \alpha_{i+}^{(t)} + \Delta \alpha_{i-}^{(t)}|$} presumes that $H_i$ is highly correlated with $\bm{x}^{(t)}$.

\textbf{Preserved information} measures how much information in a hidden state cluster $H_i$ has been retained after processing a new input $\bm{x}_t$. \revise{The preserved information of $H_i$, $\beta_i^{(t)}$, is defined as:
\begin{equation}
\beta_i^{(t)} = \sum_{h_j \in H_i} | h_j^{(t-1)} | {\min ( 1, \max ( 0, \frac{ h_j^{(t)} }{ h_j^{(t-1)} } )) }
\label{eq:clip}
\end{equation}
where the latter min-max term is used to clip the value of $h_j^{(t)} / h_j^{(t-1)}$ into range $(0,1)$. $\beta_i^{(t)}$ can be thought as the intersection volume between current and previous aggregate information $\alpha_i^{(t)}$, $\alpha_i^{(t-1)}$.}
For LSTM or GRU, gate information can be directly used instead of the zero to one clipper in \autoref{eq:clip}.
This index is useful for examining the overall hidden state updating characteristics. 


\section{Visualization Design} \label{sec:design}
In this section, we discuss the design choices and interaction designs of RNNVis based on the design requirements (\autoref{sec:requirement}).

\subsection{Visual Representation}
Hidden state clusters and word clusters have different intrinsic characteristics (i.e., one is abstract components while the other is interpretable texts). Thus, to present a better overview of RNN models (\autoref{req:overview}), we visualize word clusters as word clouds\revise{\cite{forbes2012visualizing}} and hidden state clusters as memory chips. Together with the \revise{sequence-level} representation, we organize the main view with a three-part layout, as shown in \autoref{fig:teaser}. PivotPath \cite{pivotpath12}, ConVis \cite{convis14} and NameClarifier \cite{nameclarifier17} have inspired our layout design, which reduces visual \revise{clutter} across different entities and provides an easy interface for interaction design.

\subsubsection{Hidden State Clusters as Memory Chips}
\revise{We visualize each hidden state unit as a small \revise{square-shaped} memory cell and \revise{pack} memory cells in the same cluster into a rectangular memory chip to allow exploration of details (\autoref{req:detail}). All the memory chips are vertically stacked and aligned by their centers (\autoref{fig:teaser}C) to present an overview of RNN's hidden state mechanisms (\autoref{req:overview}).}
By default, memory chips have the same heights so that larger hidden state clusters will be visualized with larger width. In case of unbalanced clustering results, we also provide an alternative layout in which memory chips are aligned with equal widths to achieve a space efficient arrangement. This design is inspired by the association between RNN and Turing machine, where hidden state units can be associated \revise{with} memory cells, which is intuitive for computer science researchers. 

With the memory-like layout, we also use a divergent blue to orange color to represent the response value of hidden state units based on interaction. For example, with a given word with embedding $\bm{x}$ selected, \revise{each hidden unit $h_j$ is rendered with the value of estimated expected response $s(\bm{x})_j$. }

Since our experiments show that a hidden state cluster of a typical RNN can contain up to a few hundred units, we adaptively pack memory cells into a few rows according to the size of hidden state cluster and the current window size for scalable visualization. The current implementation can handle up to 20 clusters with two thousand hidden state units.

\subsubsection{Word Clusters as Word Clouds}\label{sec:design-word}
Each word cluster is visualized as a word cloud, and all the word clouds are arranged in a circular layout as shown in \autoref{fig:teaser}D. To utilize the space more efficiently, word clouds are bounded by rectangles rather than circles to show more words. Word clouds present summary interpretations (\autoref{req:interpret}) of hidden state clusters, and allow users to navigate through clusters to validate the clustering quality. When users click on a hidden state cluster, highly correlated word clouds are highlighted to present users with an overall interpretation of the cluster. Word clouds also serve as an entry interface for analyzing an RNN's detailed behavior on individual words. 

In a word cluster \revise{$W_i$}, the size of each word $w_{ij}$ is proportional to the \revise{Euclidean} distance $d(c_i, w_{ij})$ between $w_{ij}$ and the cluster's centroid $c_i = \sum_{j} w_{ij} / |W_i|$. Thus, more centered words will have larger sizes in the word cloud. This design allows users to quickly identify the most representative words of a cluster. To maintain the completeness of the interpretation rather than only showing a few representing words, \revise{we linearly scaled} the words to readable size. 

Users can also turn on the grammar mode, which rendered words into different colors according to their part of speech (POS) tags. For words that have multiple possible POS tags, we render them according to its most frequent POS tag in the datasets. For implementation details, we use a pre-trained Greedy Averaged Perceptron tagger in NLTK \cite{nltk09} to first tag words with the PTB tag set, then convert the POS tags into universal tag set \cite{lrec12pos} which contains only 12 tags. 

\subsubsection{Co-Clustering Layout}
As shown in \autoref{fig:teaser}, the memory chips and word clouds are separately positioned and linked (\autoref{req:overview}). To avoid visual clutter, we aggregate the bipartite connections between words and hidden state units into cluster-to-cluster edges, which is a similar practice of Xu et al. \cite{xu16pvis}. The width of a cluster-to-cluster edge indicates the aggregate correlation between a word cluster $W_i$ and a hidden state cluster $H_j$, which is computed as \revise{the average of bi-connected edge weights}:
\begin{equation}
\revise{e(W_i, H_j)} = \frac{1} {|W_i| \times |H_j|} \sum_{w_x \in W_i, h_y \in H_j} \revise{s(\bm{x}(w_x))_y}
\end{equation}
where $|W_i|$ and $|H_j|$ are cardinalities of $W_i$ and \revise{$H_j$}.
To distinguish the positive and negative correlations, we use blue and orange to represent negative and positive edge weights. The blue-orange diverging color scheme is consistently used to visualize negative and positive values throughout our design.

With spectral co-clustering, hidden states and words are clustered into one-to-one clusters. Thus word clouds can be easily positioned with the same order as hidden state clusters to minimize visual clutter. When using other co-clustering algorithms, where co-clusters does not form pairs, \revise{force-based layouts} can be applied to position word clouds.

Here the circular layout of word clouds is designed to reduce visual clutter of the edges between bipartite clusters \cite{convis14}. Moreover, the circular arrangement is cleaner and more time efficient than a pure force-based layout as proposed by Xu et. al \cite{xu16pvis}. 

\begin{figure}[ht]
 \centering 
 \includegraphics[width=\columnwidth]{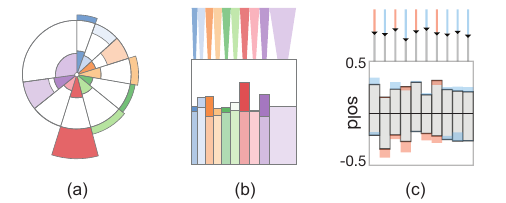}
 \caption{Alternatives of the glyph design for sequence node. (a) a pie chart based design. (b) a bar chart design with bar width encoding cluster size and (c) the composite bar chart glyph used in RNNVis.}
 \label{fig:glyph-design}
\end{figure}

\subsubsection{Glyph Design of Sequence Nodes}\label{sec:design-seq}
To enable sequence-level analysis of RNN (\autoref{req:sequence}), we design a glyph-based sequence visualization. Here a sequence refers to a sentence or a paragraph, and the nodes denote words of the sentence. The measurements used in the glyph design are discussed in \autoref{sec:model-sequence}. 

As shown in \autoref{fig:glyph-design}(c), the bar chart based glyph is designed to help experts \revise{understand} the sequence-level behavior of an RNN. The bold rectangle bounded part of each bar denotes the average aggregate information $\alpha_i^{(t)} / |H_i|$ of a hidden state cluster $H_i$, split by the horizontal zero line into positive information $\alpha_{i+}^{(t)} / |H_i|$ (upper part) and negative information $\alpha_{i-}^{(t)} / |H_i|$ (lower part). The top and bottom colored ``hat'' of each bar represents the updated information $\Delta \alpha_{i+}^{(t)} / |H_i|$ and $\Delta \alpha_{i-}^{(t)} / |H_i|$ respectively. To keep a consistent design style, we use orange color to encode an increase in the positive information or a decrease in the negative information, and use blue color to encode the opposite. Note that the increased information is always a part of the aggregate information, while the decreased information is placed outside the bounding box. The order of the bar \revise{is} consistent with the order of hidden state clusters in the co-cluster layout. Users can also hover on any bar to highlight its corresponding hidden state cluster. At the top of the squared glyph is a control chart showing the percentage of information that flows from \revise{the} previous step. The position of the cursor on each vertical line represents the preserved information normalized by the corresponding aggregate information $\beta_i^{(t)} / (\alpha_{i+}^{(t)} - \alpha_{i-}^{(t)})$ in a hidden state cluster.

During the design process, we have considered several design alternatives as shown in \autoref{fig:glyph-design}. Although a pie chart based glyph (\autoref{fig:glyph-design}(a)) is aesthetic, it cannot be used to compare the aggregate information and updated information simultaneously. As for the variable-width bar chart (\autoref{fig:glyph-design}(b)) which encodes cluster sizes as bar widths, the experts found it inconvenient to identify small clusters. Also, the choices of categorical colors are limited when the number of clusters is large. Comparing all three glyph designs, the experts felt design (c) is the most helpful one, since they can easily tell the direction (positive or negative) and extent of the information change in each cluster by simply identifying the major color of the corresponding bar.

Based on the glyph design, a sentence is visualized as a sequence of glyph nodes, as shown in \autoref{fig:teaser}(B). Users can click on a single node to highlight the links to its most responsive hidden state clusters. Here the \revise{link width} between a node $w_t$ and hidden state cluster $H_i$ is proportional to the absolute value of average updated information $|\Delta \alpha_{i+}^{(t)} + \Delta \alpha_{i-}^{(t)}|/ |H_i|$. The color of the link denotes the updated information $\Delta \alpha_{i+}^{(t)} + \Delta \alpha_{i-}^{(t)}$ in this cluster is positive (orange) or negative (blue).

\begin{figure}[ht]
 \centering 
 \includegraphics[width=\columnwidth]{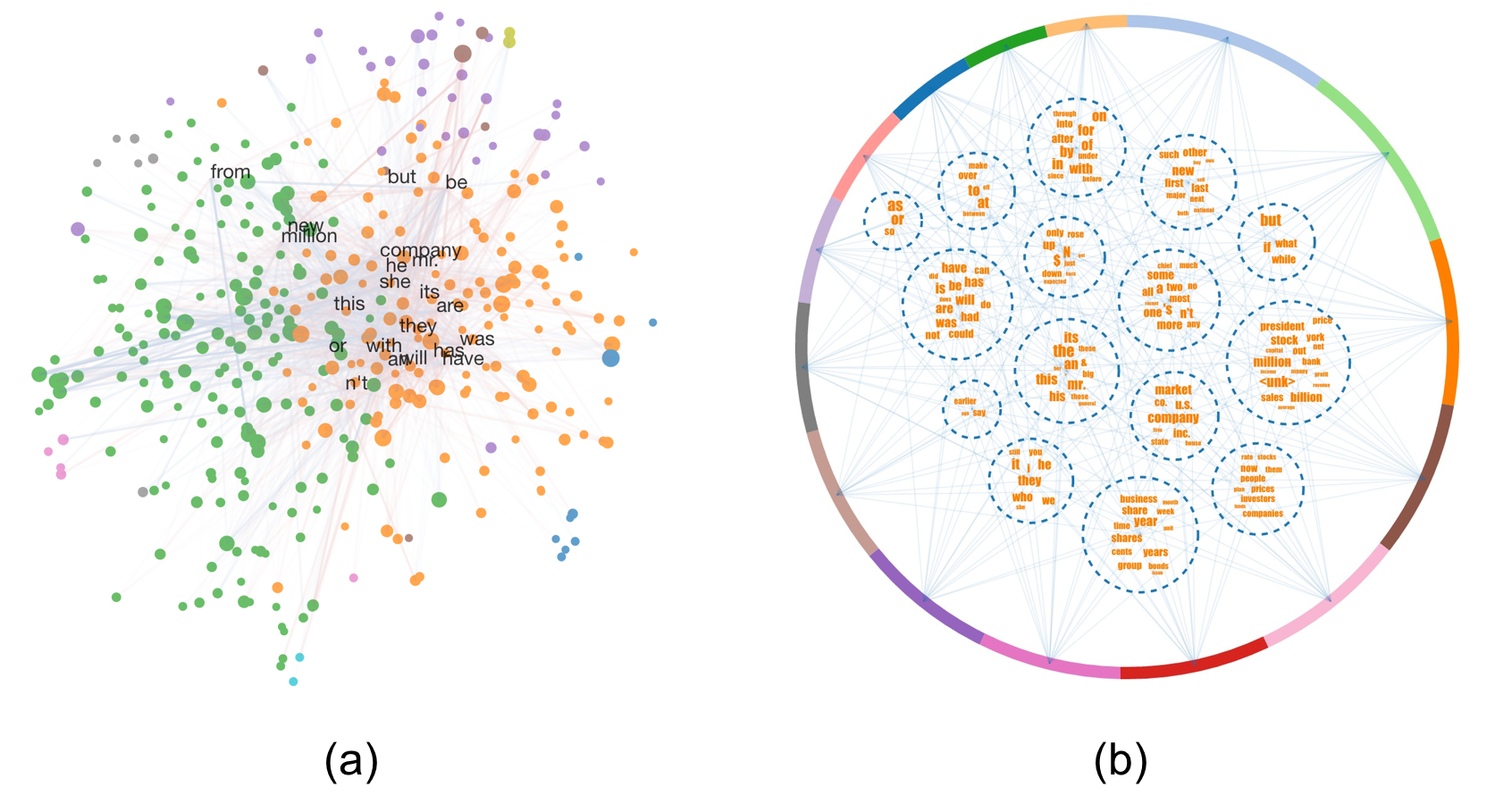}
 \caption{Alternative layouts of the main view. (a) a design that combines t-sne projection of hidden states with force-directed layout of words. (b) a design that visualizes hidden state clusters as arcs and word clusters as word clouds.}
 \label{fig:alternative-layouts}
\end{figure}

\subsubsection{Layout Alternatives}

Several alternative designs were considered before the co-cluster based bipartite visualization was finalized.

As shown in \autoref{fig:alternative-layouts}(a), an alternative layout that we proposed at the beginning uses t-SNE to project and fix hidden state units in 2D space and then arranges the positions of words using force simulation. Though this design shows hidden states projections and words in the same space, this design is not scalable for \revise{a} large number of states or words where details are messed with numerous links.

Another alternative design visualizes word clusters as word clouds and positions them inside the ring with force-directed layout (\autoref{fig:alternative-layouts}(b)). This design handles bi-connections between hidden states and words better than the previous one, but it is difficult to use in general comparative analysis (\autoref{req:compare}). The experts also felt difficult to associate the circular arcs with hidden states, which introduces confusion to users.

\subsection{Interactions}\label{sec:interaction}
To assist user's exploration, we develop rich interactive features through \revise{the} main view to \revise{the} detailed view for RNNVis.

\subsubsection{Detail on Demand}
To avoid overloading users with too much information, all interaction designs are guided by visual information seeking mantra: ``overview first, zoom and filter, then details-on-demand'' \cite{shneiderman96}.

When users select a hidden state cluster, edges to its highly correlated word clusters are highlighted. Meanwhile, these word clouds are expanded to show the complete word collections. Other word clouds are contracted to only display a few representative words.

Except for cluster-level interaction, detailed statistics of individual hidden state units and words are displayed in the detail view upon interaction. When users select a hidden state unit, the activation distribution of a hidden state unit on different words will be displayed as box plots, as shown in \autoref{fig:teaser}(E). Users are also allowed to click on individual word $w$ in the word cloud and see the model's expected response $s(\bm{x}(w))$ visualized as a heat map in the memory chips visualization.

In sequence analysis, users are allowed to zoom in and out certain parts of the sequence with brushing. Similar to the co-cluster layout, the links between a sequence node and the memory chips are highlighted when users click on the node.

\begin{figure*}[bht]
 \centering 
 \includegraphics[width=1\textwidth]{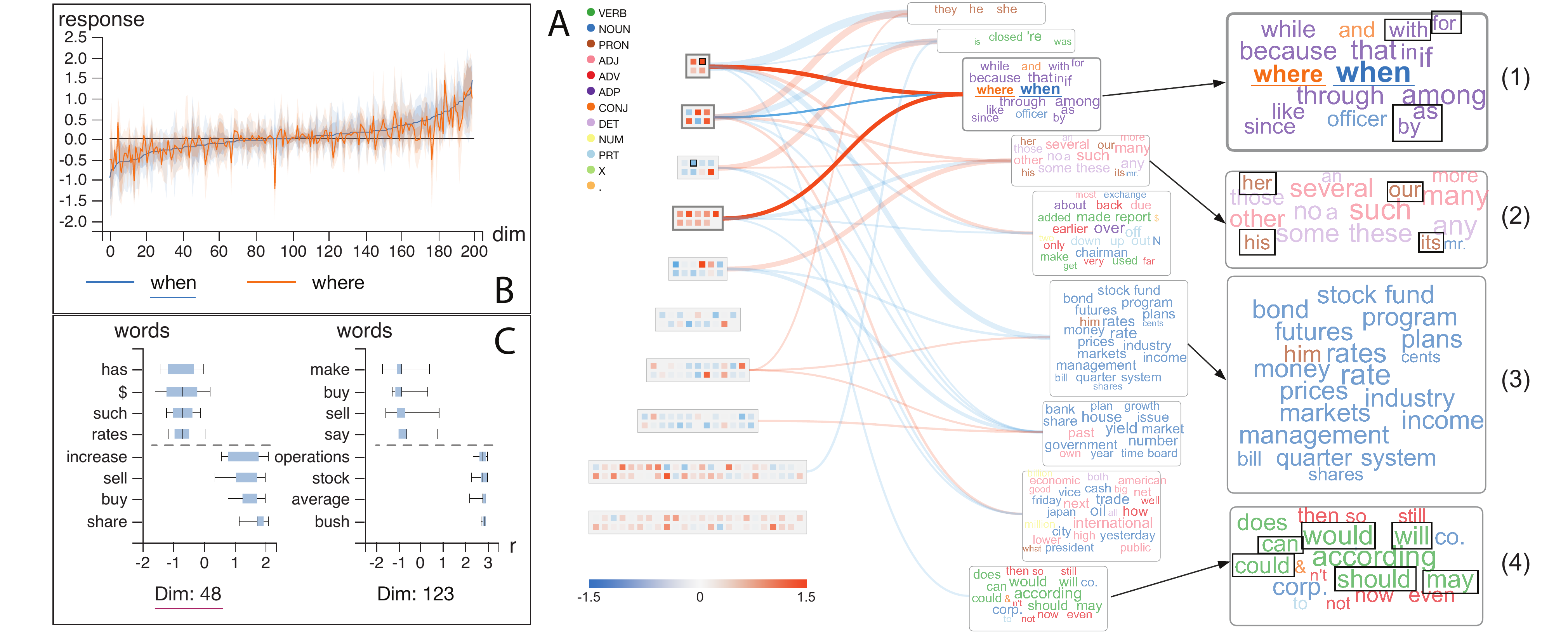}
 \caption{Co-cluster visualization of the last layer of LSTM-Small. (A): The co-Clustering layout with words colored according to POS tags. (B): The detail view showing model's responses to ``when'' and ``where'' in overlay manner. (C): Words interpretations of the 48th and 123rd cell states.}
 \label{fig:eval-introduce}
\end{figure*}

\subsubsection{Comparative Interactions}
RNNVis uses three levels of comparison to help experts understand RNNs (\autoref{req:compare}), namely, hidden state unit and \revise{word-level} comparison, sentence-sentence comparison and model-model comparison.

For \revise{word-level} comparison, users can interactively select two or three words in the word clouds, and compare the distributions of model's responses to these words in an overlay manner. Since a hidden state vector may have hundreds of dimensions, the dimensions can be sorted according to the expected response to a selected word $s(\bm{x}(w))_j$ for a clearer comparison. For example, as shown in \autoref{fig:he-she-for}, the dimensions are sorted according to the word ``he''. We can see that ``he'' and ``she'' result in a similar response distribution, while the response distribution of ``for'' is very different from ``he''. This condition suggests that the RNN model has learned to distinguish pronouns and prepositions. Similar comparisons can be performed for hidden states using a side-by-side box plots layout. 

To compare how an RNN treats different sentences, we provide a \revise{side-by-side} comparison as shown in \autoref{fig:eval-sentiment}. Users can easily analyze whether different sentence contexts may dramatically influence the information inside the hidden states.

As shown in \autoref{fig:compare-models}, a side-by-side layout of the main view is employed to compare different models.

\section{Evaluation}
In this section, we first demonstrate how RNNVis can be effectively used to help experts understand and explore the behavior of different RNN models through two case studies. Then we present the feedback gathered through one-to-one interviews with the domain experts.

\begin{table}[bth]
  \caption{Configuration and performance of RNN models for language modeling (lower perplexity represents better performance).}
  \label{tab:model-config}
	\centering%
    \begin{tabular} {c c c c c}
    \hline
    \multicolumn{3}{c}{Model} & \multicolumn{2}{c}{Perplexity} \\ 
    Model & Layer & Size & Validation Set & Test Set \\
    \hline
    LSTM-Small & 2 & 200 & 118.6 & 115.7 \\
    LSTM-Medium & 2 & 600 & 96.2 & 93.5 \\
    LSTM-Large & 2 & 1500 & 91.3 & 88.0 \\
    RNN & 2 & 600 & 123.3 & 119.9 \\
    GRU & 2 & 600 & 119.1 & 116.4 \\
    \hline
    \end{tabular}
\end{table}
\subsection{Case Study: Language Modeling}
The first case study is collaborated with expert E1 to understand and compare different RNN models that were trained for language modeling, which is a basis of machine translation and image captioning. Note that for simplicity, the expected response \revise{$s(\bm{x})$} that we used in this case study are estimated on hidden state observations in the test set, though the training set or the validation set may also be evaluated.

A \textbf{language model} assigns probabilities to sequences. Its target is to predict the conditional probability of the next word given the sequence of previous words $P(w^{(t)} \mid w^{(0)} \cdots w^{(t-1)})$. In this case study, we used the Penn Tree Bank (PTB) dataset \cite{ptb93}, which is a widely used benchmark dataset for language modeling. 
We trained a vanilla RNN, a GRU, and three LSTMs on the PTB dataset. The detailed parameter settings and performance of these models are shown in \autoref{tab:model-config}. 

\textbf{An ablation of an LSTM.}
We started with \revise{an LSTM model} (LSTM-Small in \autoref{tab:model-config}) to introduce the basic usage of RNNVis to expert E1. The number of \revise{clusters} was set to 10, and the bipartite links with absolute weight less than 20\% of the maximum value were filtered. The visualization shows the cell states of the last layer.

At the first glance \revise{(\autoref{fig:eval-introduce}A)}, the expert noticed that words with similar functions were likely to be clustered together. \revise{For example, prepositions like ``with'', ``for'', ``by'' and ``as'' were grouped into the word cloud (1), and modal verbs like ``would'', ``could'' ``should'' and ``will'' were clustered to the word cloud (4). After focusing on the word cloud (1), the first and fourth memory chips are highlighted. As mentioned in \autoref{sec:design-word}, word clouds visualize a batch of words as a summary interpretation of hidden state clusters. The hidden units in the first and fourth memory chips can be regarded to be able to capture the information of prepositions (\autoref{req:overview}). The expert then turned on the POS mode to color words according to their most frequent POS tags to further evaluate the quality of the word clusters. The result clearly showed that LSTM-Small was able to distinguish a large number of nouns and prepositions from other words. Although the word cloud (2)} was messed with adjectives, determinants, and pronouns (especially adjectival possessive pronouns like ``her'', ``his'', ``our'' and ``its''), these words actually contain similar decorative semantics for specifying the successive nouns.
\begin{figure*}[bth]
\includegraphics[width=1\textwidth]{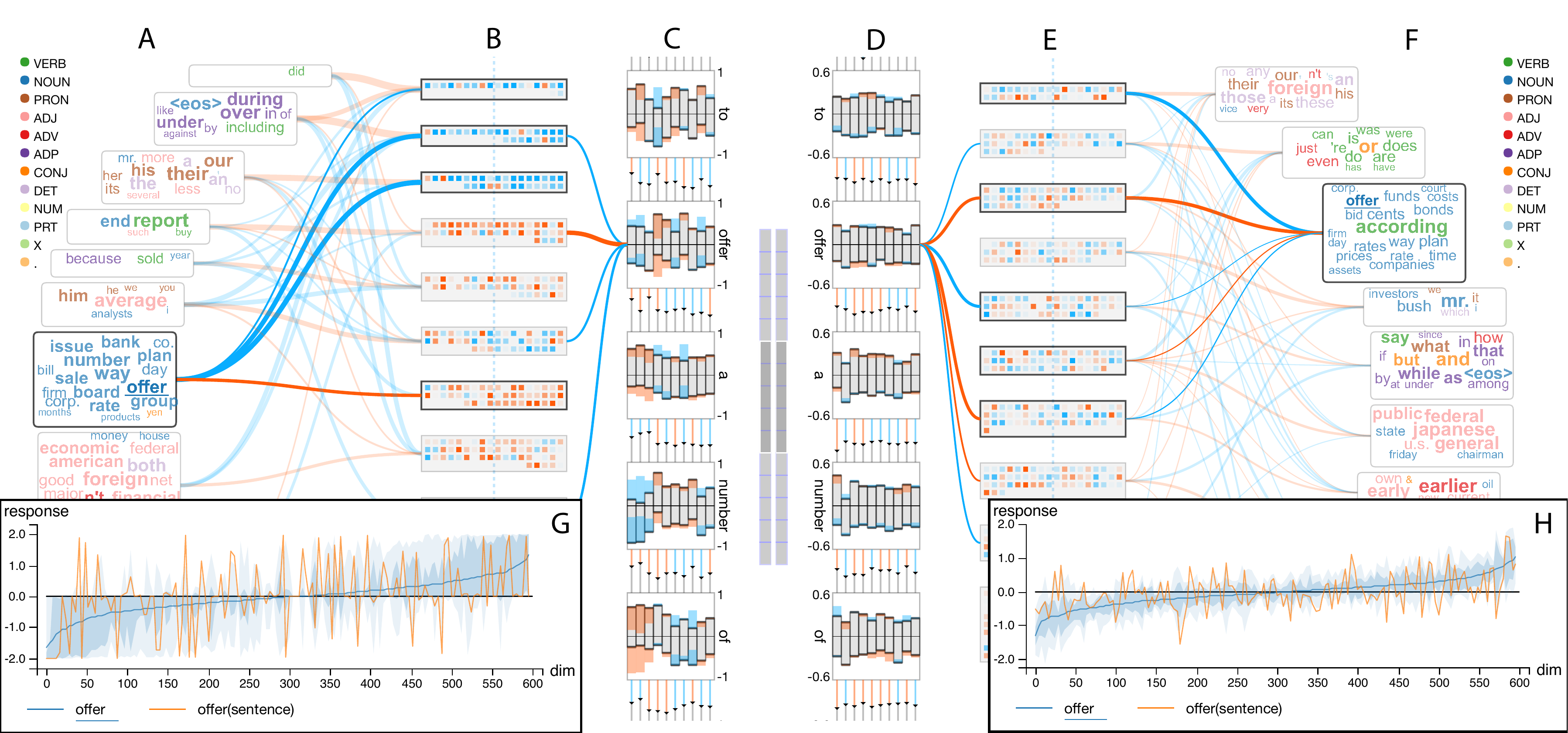}
\caption{Comparison of an RNN and an LSTM. Left \revise{(A-C)}: co-cluster visualization of the last layer of RNN (\autoref{tab:model-config}). Right \revise{(D-F)}: visualization of the cell states of the last layer of LSTM-Medium. Bottom (\revise{GH}): two models' responses to the same word ``offer''. The orange line represents the actual response in the sentence.}
\label{fig:compare-models}
\end{figure*}

To further validate his hypothesis on the model's ability of recognizing POS tags, the expert clicked on several pairs of words (e.g., ``those'' and ``these'', \revise{and ``his'' and ``her''}) and compared the detailed distribution of model's responses \revise{shown in \autoref{fig:eval-introduce}B ( \autoref{req:detail})}. The results showed \revise{that the distributions of words in each pair are only slightly different from each other, indicating the model can capture their similar linguistic functions}.
Then the expert explored among detailed interpretations of specific hidden units by selecting his interested cells on the memory chips. The box plots in the detail view (\autoref{fig:eval-introduce}C) showed that both the 123rd and 48th dimensions can catch information of verbs like ``buy'' and ``sell'', although their information is stored with opposite signs in cell states.
\revise{Since} the estimated expected responses \revise{$\hat{s}(\bm{x})$} were directly computed from hidden states observations, the similarity between word pairs was a result of model's hidden state updating mechanism rather than similarity in word embedding alone. This \revise{suggests} that LSTM-Small can well recognize grammatic functions of words, although it only has 200 cell states, and it was not intentionally trained for POS tagging. 


\textbf{Comparison between vanilla RNN and Gated RNNs.}
Despite the interesting findings on the interpretable representations of LSTMs, the expert was also curious \revise{about} whether these visualizations can reveal the difference between the vanilla RNN and other gated RNNs (e.g., LSTM and GRU). 

The expert first compared an RNN with an LSTM (LSTM-Medium) using the compare mode of the main view (see \autoref{fig:compare-models}, \autoref{req:compare}). \revise{The memory of the RNN model (B) has significantly higher saturation than that of the LSTM (E). This indicates the expected response strength of the RNN tends to be much stronger. The expert then added a sentence: ``The company said it planned to offer a number of common shares in exchange,'' for the two models to run and compare their response histories (\autoref{fig:compare-models}C and D). The colored part of each bar chart glyph of C is much larger than that of D, indicating the RNN updates its hidden memories more intensively than the LSTM.  Or, the LSTM's response to inputs is generally much sparser than the RNN (see \autoref{fig:compare-models}).} LSTM's lazy behavior in updating cell states might contribute to its long-term memory advantage over RNN. The idea was further validated on detailed model response distribution in the detail view, \revise{where the response distribution of the RNN (G) is more dispersive and unstable than that of the LSTM (H).}
Such evidence also provided empirical explanations of \revise{the widely accepted claim that LSTMs can handle long-term dependencies better than vanilla RNNs}.

The expert also compared GRU with LSTM and RNN. Although the response distribution of the GRU model is dispersive, the shape and range of the distribution of hidden states' response \revise{are} closer to those of the LSTM. This finding may serve as an empirical explanation of the similar performances of GRUs and LSTMs, which agrees with a recent empirical result \cite{DBLP:conf/icml/JozefowiczZS15}.




\subsection{Case Study: Sentiment Analysis}\label{sec:case-sentiment}
Besides language modeling, we also conducted another case study to understand RNN's behavior in sentiment analysis with expert E2. The dataset we used is retrieved from Yelp Data Challenge \footnote{http://www.yelp.com/dataset\_challenge}. The Yelp dataset contains over 4 million reviews for restaurants with labels from one to five. For simplicity, we pre-process the five labels into only two labels by mapping one and two to ``negative'' and four and five to ``positive''. The data entries with label three are not included. \revise{For simplicity, we only used a small subset of 20 thousands reviews with length less than 100 words with 0-1 labels. The small dataset was split into train/validation/test set with 80/10/10 percentages. We trained a single layered GRU with 50 cell states on the train set and achieved accuracies of 89.5\% and \revise{88.6\%} on the validation and test set. }

\begin{figure}[t]
 \centering 
 \includegraphics[width=\columnwidth]{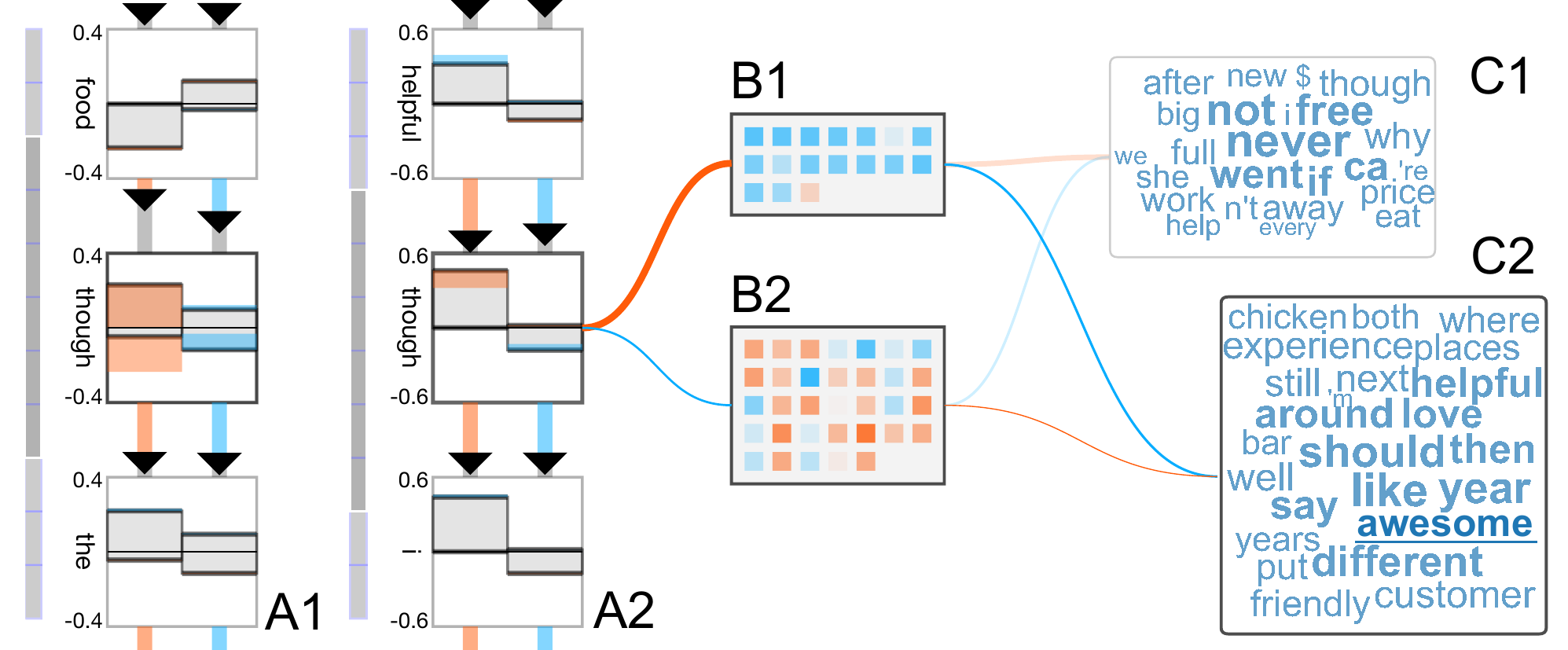}
 \caption{Comparison of two sentences on an GRU. \revise{A1 and A2: two evaluated sentences, focusing on the word ``though''. B1 and B2: hidden unit clusters. C1 and C2: word clusters. B1 is positively related to negative words, and B2 is positively related to positive words.} }
 \label{fig:eval-sentiment}
\end{figure}

\textbf{Sentiments of words.}
The expert first explored how the trained GRU handled the sentiment of each word. Expecting to have hidden state clusters that respectively catch positive, negative information, the expert set the cluster number to two. \revise{As shown in \autoref{fig:eval-sentiment}, the co-clustering visualization clearly represented two word clouds that present different sentiments.} For example, the upper word cloud contained negative words such as ``never'' and ``not'', and the lower word cloud contained positive ones such as ``love'', ``awesome'' and ``helpful''. \revise{From the color of the link, we can infer that the hidden units in the upper memory chip are positively related to negative words, while the lower memory chip reacts positively to positive words. Note that such relation is coupled, i.e., hidden states use positive and negative values to encode opposite sentiments.}

\textbf{Sentiments in contexts.}
The expert was also very interested in how the GRU dealt with subtle sentiment changes in a sentence (\autoref{req:sequence}). Then the expert used the sentence-level comparison visualization to investigate how different context may influence the sentiment of the same word. The expert compared two sentences: ``I love the food, though the staff is not helpful'' and ``The staff is not helpful, though I love the food''. As shown in \autoref{fig:eval-sentiment}, the updated information of the first hidden state cluster of ``though'' in sentence (a) is much larger than in sentence (b), denoting a larger response towards negative sentiment. Note the previous information of the first sentence (at word ``food'') has more positive sentiment than that of the second sentence (at word ``helpful''). These results suggest that the GRU had learned to treat the sentiment of the same word differently according to different contexts.

\begin{figure}[t]
 \centering 
 \includegraphics[width=\columnwidth]{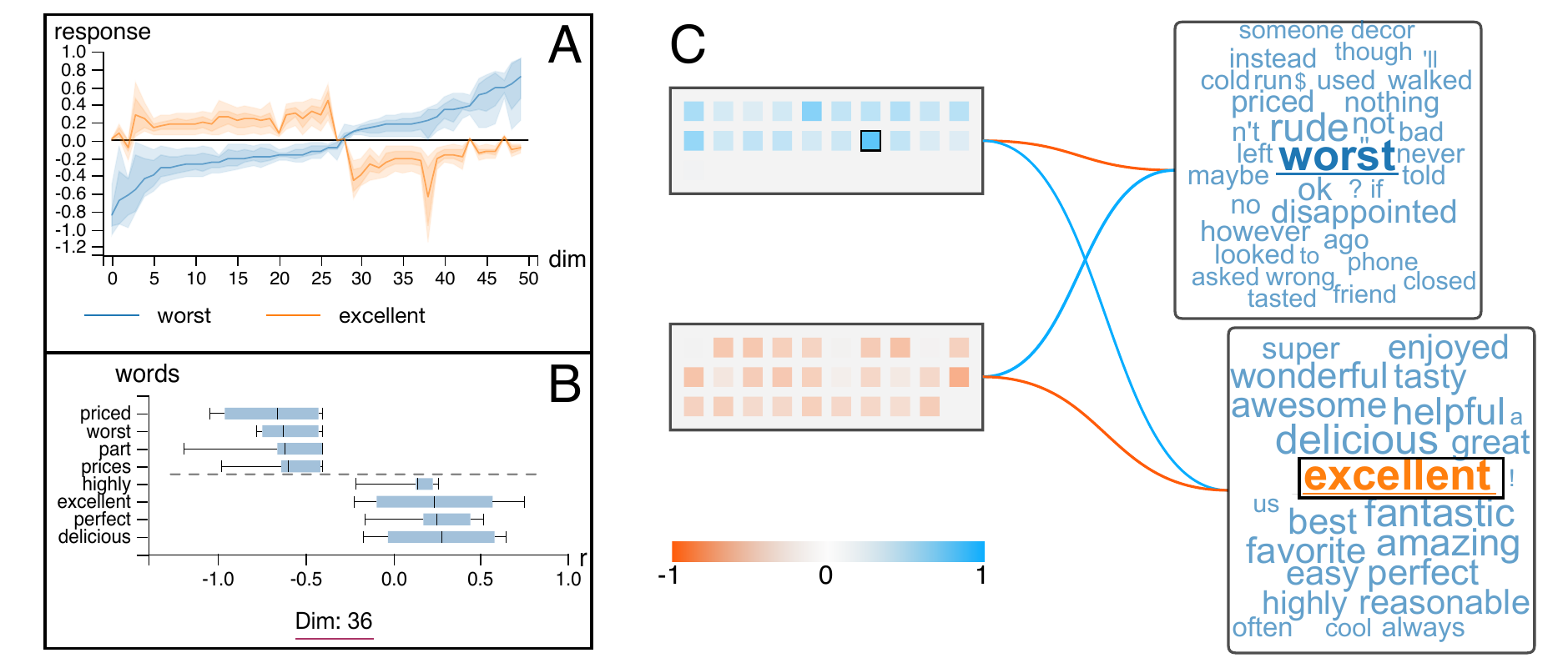}
 \caption{ \revise{The GRU model trained on the balanced dataset. A: the model's responses to word ``worst'' and ``excellent''. B: most responsive words of the 36th hidden unit. C: co-cluster visualization of the model.} }
 \label{fig:eval-sentiment2}
\end{figure}

\revise{\textbf{Diagnosing sentiment models.} Though the visualization in \autoref{fig:eval-sentiment} indicates the GRU model is capable of distinguishing different sentiments, many negative words such as ``bad'' and ``worst'', which can be found in the dataset, are not shown in the visualization. The expert suspected that the model might have different performance in positive and negative reviews. We then examined the code base and found that the dataset was unbalanced, with approximately 3:1 positive to negative reviews, which caused the uneven performance of the model. After over-sampling to achieve a balanced dataset, a re-trained GRU model achieved accuracies of 91.52\% and 91.91\% on the validation set and test set. The visualization of the new model (\autoref{fig:eval-sentiment2}) shows more words with strong sentiments, such as ``rude'', ``worst'' and ``bad'' in the negative word cloud, and ``excellent'', ``delicious'' and ``great'' in the positive word cloud. By clicking words ``worst'' and ``excellent'', the expert compared the response distributions of the two words in the detail view (A). We can see that they have nearly opposite response patterns (i.e., the dimensions with negative response to ``worst'' always respond positively to ``excellent'').
The expert commented, ``the visualization provides clues on what's going on inside the model'' and ``allows me to explore the model's behavior from different perspectives''.
}

\subsection{Expert Reviews}
\revise{
To evaluate the effectiveness and usefulness of our system, we conducted one-to-one interviews with four domain experts (P1, P2, P3 and P4). P1 and P2 are machine learning researchers who have published papers on prestigious conferences in related fields. P1 specializes in sentiment analysis while P2 is interested in NLP and transfer learning. P3 and P4 are graduate students with knowledge in both machine learning and visualization. We first introduce the proposed system and the used method through the tour guide and the presentation of the first case study. Then we asked the experts to explore and compare the two GRU models discussed in \autoref{sec:case-sentiment}. Finally, we asked them to guess which model has better performance and state their findings and suggestions. The result and feedback are summarized as follows.
}

\revise{
\textbf{Effectiveness.} After exploring the two models using RNNVis, all the interviewees made correct guess on the performances of the models. They can also clearly state that the second model can capture more obvious negative words than the first one, which indicates the second model is more likely to provide better predictions on the sentiments of reviews. Besides, they all agreed on the effectiveness and usability of the system. P2 commented that given the success of neural models, a recent concern in the community is their interpretability. RNNVis provides him with the possibility to flexibly explore RNNs' hidden behavior, and easily validate his hypothesis on these models. The comparison of vanilla RNNs and LSTMs reveals the long-term dependency characteristics of LSTMs intuitively and convincingly. P1 added that this tool could help her make more sense on the model: ``it provides intuitive information on what is happening in sentiment models, which is helpful for debugging compared with merely looking at training logs''. P3 mentioned that ``tuning deep learning models is like metaphysics to him, and this visualization can be very helpful when introducing the advantages of RNNs to non-experts''. 
}

\revise{
\textbf{Cognitive Burdens.} All the experts agreed that they could fully understand the visual encoding of the system after our introduction and the guidance tour. They agreed that the cognitive load is acceptable and the interaction is fluent. P4 commented that the visualization design is quite aesthetic and he enjoys using it for exploration. However, at the beginning of the interview, P2 has mistakenly regarded different memory chips as different layers of the  network. He also suggested us to design a simple document for users to learn the techniques behind the system quickly. 
}

\revise{
\textbf{Suggestions.} The experts have also provided valuable suggestions on improving our work. P1 and P3 suggested that we make the system programmable so that other visualizations such as loss curves and gradient distributions can be shown during the training process. P4 also suggested we should add model's inference results during the sequence visualization as additional information. P2 mentioned the need for a simpler visualization for explaining RNNs' behavior during presentation. P1 and P2 pointed out that we can explore more applications such as machine reading and more advanced RNN-based models such as attention mechanisms and memory networks.
}

\section{Conclusion}
In this paper, we presented a visual analytic method for understanding and diagnosing RNNs \revise{for text applications}. A technique based on RNN's expected response to inputs is proposed to interpret \revise{the} information stored in hidden states. Based on this technique, we designed and implemented an interactive co-clustering visualization composed of memory chips and word clouds, which allows domain users to explore, understand and compare the internal behavior of different RNN models. Our evaluation, including two case studies and expert interviews, demonstrated the effectiveness of using our system to understand and compare different RNN models and also verified the completeness of design requirements.

To further improve our proposed visual analytic system, RNNVis, we plan to deploy it online, and improve the usability by adding more quantitative measurements of RNN models. \revise{Considering the proposed technique is based on the discrete input space of texts, RNNVis is only \revise{suitable} to analyze RNN-based models for texts. The analysis of RNN models for audio applications requires further efforts to construct interpretable representations.} A current bottleneck for RNNVis is the efficiency and quality of co-clustering, which may results in delays during interaction. Other potential future work includes the extension of our system to support the visualization of specialized RNN-based models, such as memory networks or attention models. 

\acknowledgments{
The authors wish to thank anonymous reviewers for their constructive comments. The authors also wish to thank the domain experts who participated in the studies and Miss Yuehan Liu who helped record the video. This work was supported in part by the National 973 Program of China (2014CB340304).}

\clearpage

\bibliographystyle{abbrv}

\bibliography{template}

\clearpage
\begin{appendices}

\section{Model Definitions} \label{app:model}

\subsection{Long Short-Term Memory}
LSTM was designed by Hochreiter \& Schmidhuber \cite{hochreiter97lstm} to deal with the training difficulties of RNN. Except for hidden state $\bm{h}^{(t)}$, an LSTM maintains another memory vector $\bm{c}^{(t)}$ called \textit{cell \revise{state}}. Also, LSTM uses input gate $\bm{i}^{(t)}$, forget gate $\bm{f}^{(t)}$ and output gate $\bm{o}^{(t)}$ to explicitly control the update of $\bm{h}^{(t)}$ and $\bm{c}^{(t)}$. The three gate vectors $\bm{i}^{(t)}, \bm{f}^{(t)}, \bm{o}^{(t)} \in \mathbb{R}^n$ are computed using:
\begin{equation}
\begin{aligned} 
\bm{i}^{(t)} &= \sigma(\bm{W}_i \bm{h}^{(t-1)} + \bm{V}_i \bm{x}^{(t)}) \\
\bm{f}^{(t)} &= \sigma(\bm{W}_f \bm{h}^{(t-1)} + \bm{V}_f \bm{x}^{(t)}) \\
\bm{o}^{(t)} &= \sigma(\bm{W}_o \bm{h}^{(t-1)} + \bm{V}_o \bm{x}^{(t)})
\end{aligned} 
\end{equation}
and cell \revise{state} and hidden \revise{state} are updated by:
\begin{equation}
\begin{aligned} 
\tilde{\bm{c}}^{(t)} &= \tanh(\bm{W}_c \bm{h}^{(t-1)} + \bm{V}_c \bm{x}^{(t)}) \\
\bm{c}^{(t)} &= \bm{f}^{(t)} \odot \bm{c}^{(t-1)} + \bm{i}^{(t)} \odot \tilde{\bm{c}}^{(t)} \\
\bm{h}^{(t)} &= \bm{o}^{(t)} \odot \bm{c}^{(t)}
\end{aligned}
\end{equation}
where $\tilde{\bm{c}}^{(t)}$ is called candidate cell state, and $\bm{W}_*$ and $\bm{V}_*$ are trainable weight matrices. With \revise{the} sigmoid function, the values of $\bm{i}^{(t)}, \bm{f}^{(t)}, \bm{o}^{(t)}$ have range between 0 and 1, representing how much we update the $\bm{c}^{(t)}$, forget the $\bm{c}^{(t-1)}$ and output to $\bm{h}^{(t)}$. Such an architecture was argued to have the advantage to maintain information over a relatively long term.

\subsection{Gated Recurrent Unit}

GRU is a simpler model proposed by Cho et al. \cite{cho14gru}.
A GRU only uses a single hidden state vector $\bm{h}^{(t)}$. Behaving like a simpler variation of an LSTM, a GRU uses two gates, update gate $\bm{z}^{(t)}$ and reset gate $\bm{r}^{(t)}$ to control the update of hidden memory:
\begin{equation}
\begin{aligned} 
\bm{z}^{(t)} &= \sigma(\bm{W}_z \bm{h}^{(t-1)} + \bm{V}_z \bm{x}^{(t)}) \\
\bm{r}^{(t)} &= \sigma(\bm{W}_r \bm{h}^{(t-1)} + \bm{V}_r \bm{x}^{(t)}).
\end{aligned}
\end{equation}
Then a candidate hidden state $\tilde{\bm{h}}^{(t)}$ is computed using reset gate vector before updating hidden state:
\begin{equation}
\begin{aligned} 
\tilde{\bm{h}}^{(t)} &= \tanh(\bm{W}_h \bm{h}^{(t-1)} + \bm{V}_h ( \bm{r}^{(t)} \odot \bm{x}^{(t)})) \\
\bm{h}^{(t)} &= (1 -\bm{z}^{(t)}) \odot \bm{h}^{(t-1)} + \bm{z}^{(t)} \odot \tilde{\bm{h}}^{(t-1)}.
\end{aligned}
\end{equation}
The computation of GRU can be thought as updating the hidden states vector $\bm{h}^{(t)}$ by interpolating between previous $\bm{h}^{(t-1)}$ and a candidate $\tilde{\bm{h}}^{(t)}$ with ratio $\bm{z}^{(t)}$.

\subsection{Multi-layer Models}

To increase representation capability, the aforementioned RNN models can be easily extended to multi-layered version by simply stacking. A $k$-layered RNN associates $k$ hidden states vector $\bm{h}_1^{(t)}, \cdots, \bm{h}_{k}^{(t)}$ for each layer, where each $\bm{h}_l^{(t)} \in \mathbb{R}^n$. At step $t$, layer $l$ takes the output $\bm{h}_{l-1}^{(t)}$ of previous layer and update its hidden states vector $\bm{h}_{l}^{(t)}$ using:
\begin{equation}
\bm{h}_{l}^{(t)} = f(\bm{W}_{l} \bm{h}_{l}^{(t-1)} + \bm{V}_{l} \bm{h}_{l-1}^{(t)})
\end{equation}
where we use $\bm{h}_{0}^{(t)} = \bm{x}^{(t)}$ the model input at step $t$ for the first layer.
LSTMs and GRUs can also be extended to multi-layered models with similar procedure. But we will neglect these details considering the page limits.

\clearpage
\section{Case Study: The Language of Shakespeare}

We demonstrate how new models and new datasets can be added to RNNVis, and how the system can be used to validate hypotheses related to the behavior of hidden states. 

\textbf{Training a language model on plain texts.} RNNVis provides a pre-defined pipeline to process plain texts and build up a tokenized data set. Users can easily modify configuration files to specify the vocabulary size of the new dataset and the hyper parameters of the new model (e.g., the type of RNN, the number of layers and the size of each layer). In this case study, we used a collection of Shakespeare's work\footnote{The dataset is downloaded from the Project Gutenberg: http://www.gutenberg.org/ebooks/100} (containing one million words) to build up a language modeling dataset with a vocabulary size of 15 thousand. The dataset is split into 80/10/10 train/valid/test sets. We trained three two-layer LSTM models with different sizes per each layer (see \autoref{tab:model-config2}). For the validation of the proposed hypotheses in the following paragraphs, we only illustrate the results on LSTM-Medium, though the presented phenomenons can be found in all three models.

\begin{table}[bth]
  \caption{Configuration and performance of LSTM models trained on the Shakespeare corpus (lower perplexity represents better performance).}
  \label{tab:model-config2}
	\centering%
    \begin{tabular} {c c c c c}
    \hline
    \multicolumn{3}{c}{Model} & \multicolumn{2}{c}{Perplexity} \\ 
    Model & Layer & Size & Validation Set & Test Set \\
    \hline
    LSTM-Small & 2 & 256 & 154.0 & 153.3 \\
    LSTM-Medium & 2 & 512 & 135.8 & 134.9 \\
    LSTM-Large & 2 & 1024 & 130.4 & 129.7 \\
    \hline
    \end{tabular}
\end{table}
\begin{figure}[h]
 \centering 
 \includegraphics[width=\columnwidth]{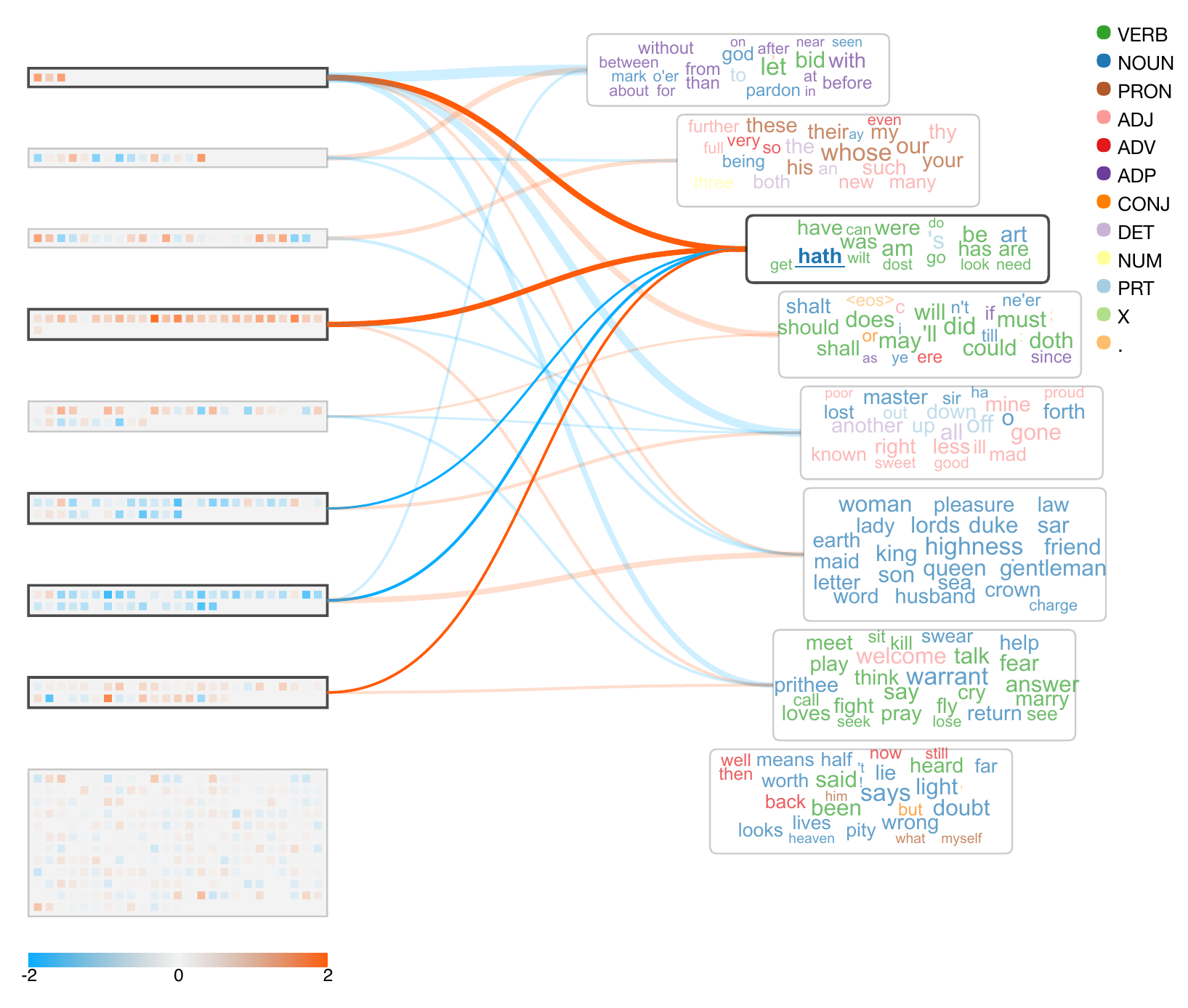}
 \caption{The LSTM-Medium model trained on the Shakespeare corpus. The POS tags of ancient words may be incorrect because the NLTK POS tagger is trained on modern English corpus.}
 \label{fig:eval-shakespeare}
\end{figure}
\begin{figure*}[ht]
 \centering 
 \includegraphics[width=2\columnwidth]{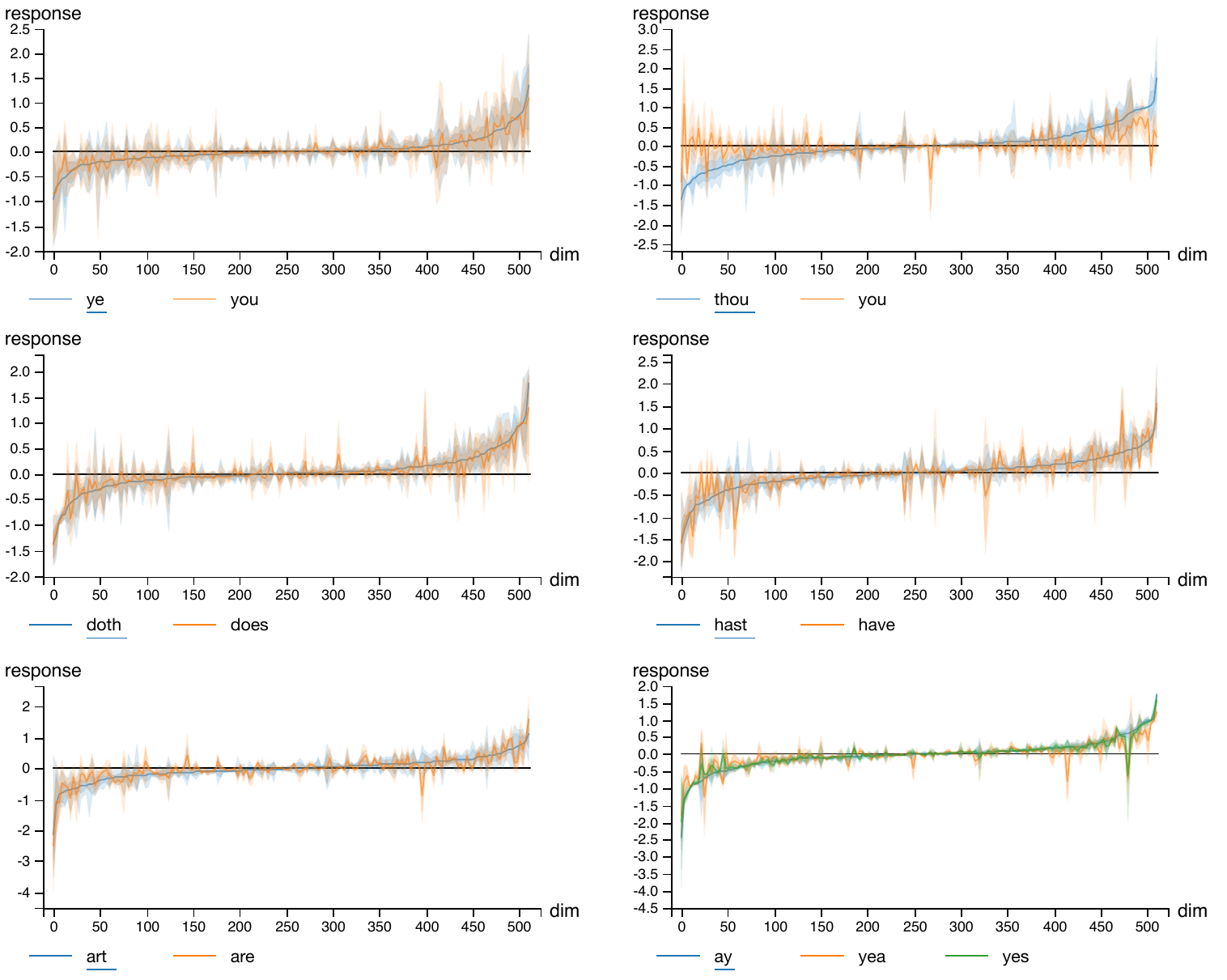}
 \caption{ The response distributions to ancient-modern word pairs of the LSTM-Medium model.}
 \label{fig:eval-shakespeare-pairs}
\end{figure*}

\textbf{A mixture of the old and the new.} Shakespeare's language is a mixture of the old and the new. For example, ``thou'' and ``you'', ``hath'' and ``has'', and ``ay'' and ``yes'' are mixed in his plays. Thus, one interesting hypothesis that we want to validate is whether the language model can capture the similar meaning of these words despite their different spellings. 

We began with the co-cluster visualization of the LSTM-Medium model (\autoref{tab:model-config2}), which is shown in \autoref{fig:eval-shakespeare}. We quickly found that ``hath'', ``dost'' and ``art'' are settled in the same word cloud with ``have'', ``do'' and ``are'', which indicates the model is able to learn the similar usage of these auxiliary verbs, no matter they are from ancient or modern English. To further validate the model's ability in learning the similarity between old and new words, we compared a variety of old-new word pairs in \autoref{fig:eval-shakespeare-pairs}. All the new words have similar response distributions as their old versions, except for ``thou'' and ``you''. After consulting an expert in classic literature, we learned that such anomaly may result from their different usages in Shakespeare's time. Although ``thou'', ``ye'' and ``you'' are all Early Modern English second person pronouns, ``thou'' is the singular form and often used in informal cases, while ``ye'' and ``you'' are plural forms and are considered to be formal and polite.

\begin{figure*}[bth]
\includegraphics[width=1\textwidth]{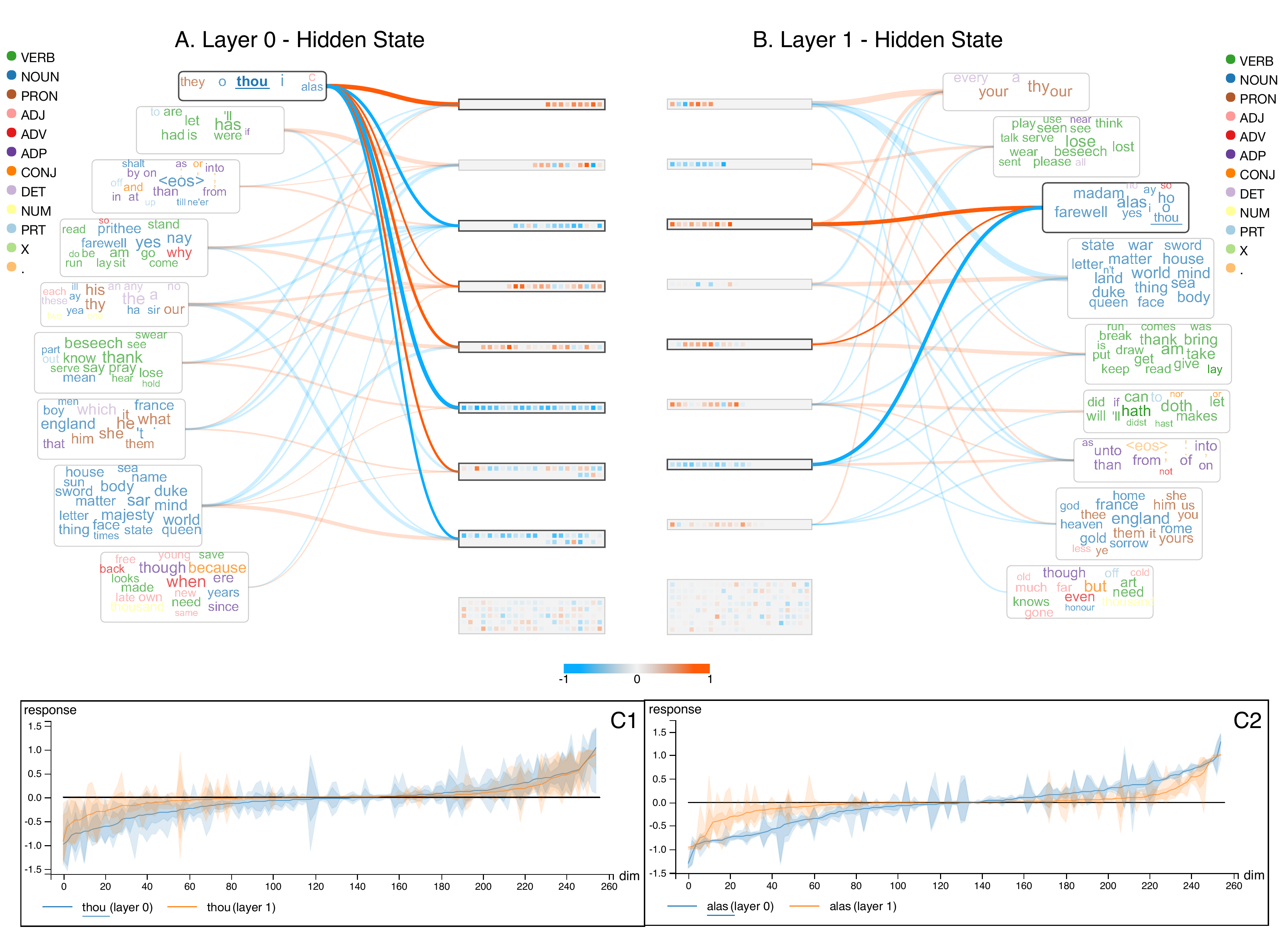}
\caption{The comparison of two layers of the LSTM model. A and B: the co-cluster visualization of the hidden states in layer 0 and layer 1, respectively. C1 and C2: the distribution of model's response to words ``thou'' and ``alas'', respectively.}
\label{fig:shakespeare-layers}
\end{figure*}

\textbf{Comparing different layers in one model.} From the visualization research of CNNs \cite{DBLP:conf/eccv/ZeilerF14}, we learned that CNNs use multiple layers to capture different levels of structure in images. Initial layers capture small features such as edges and corners, while the layers close to the output learn to identify more complex structures which formulate different classes. Little research has explored how features in RNNs are composed through multiple layers. A natural hypothesis is that the initial layers learn abstract representations, while the latter layers learn more task-specific representations. Next we illustrate how RNNVis can be used to  compare the response of different layers in the LSTM model.

Comparing the hidden state clusters of the two layers, we found that the sizes of memory chips of the layer 0 are more balanced than those of the layer 1. We also found that in layer 1, the color within each word cloud is more consistent. In another word, the quality of the word clouds in layer 1 is generally higher that those in layer 0. This indicates that the layer 0 is ``fuzzier'' in treating the usage of different words, while the layer 1 learns a ``clearer'' way in identifying the different functions of words.
Then we clicked on the same word ``thou'' of each side. As shown in \autoref{fig:shakespeare-layers}C1, the color of memory chips of layer 1 has lower saturation than that of the layer 0, indicating the hidden states of layer 1 have sparser response to the word ``thou''. Similar phenomenon could be found for other words. For language modeling, such sparsity is helpful in the last layer, where the output is directly projected through softmax to a large space containing thousands of words. To summarize, the comparative visualization of the two layers provides a way of explaining how multi-layer RNNs utilize multiple layers of hidden states to formulate the learned representations. That is, the model uses the first layer to construct abstract representations, while uses the last layer to formulate more task-specific representations.




\end{appendices}
\end{document}